\documentclass[runningheads]{llncs}
\usepackage{subfigure}
\usepackage{caption}
\usepackage{graphicx}
\usepackage{amsmath,amssymb} 
\usepackage{color}
\usepackage[width=122mm,left=12mm,paperwidth=146mm,height=193mm,top=12mm,paperheight=217mm]{geometry}
\usepackage{array}
\begin{document}
\pagestyle{headings}
\mainmatter

\title{Verisimilar Image Synthesis for Accurate Detection and Recognition of Texts in Scenes} 

\titlerunning{Verisimilar Image Synthesis for Detection and Recognition of Texts}

\authorrunning{F. Zhan, S. Lu, and C. Xue}

\author{Fangneng Zhan, Shijian Lu, and Chuhui Xue}


\institute{School of Computer Science and Engineering,\\
	Nanyang Technological University\\
	\email{ \{fnzhan,shijiian.lu\}@ntu.edu.sg},
	\email{ \{xuec0003\}@e.ntu.edu.sg}
}

\maketitle

\begin{abstract}
  The requirement of large amounts of annotated images has become one grand challenge while training deep neural network models for various visual detection and recognition tasks. This paper presents a novel image synthesis technique that aims to generate a large amount of annotated scene text images for training accurate and robust scene text detection and recognition models. The proposed technique consists of three innovative designs. First, it realizes ``semantic coherent" synthesis by embedding texts at semantically sensible regions within the background image, where the semantic coherence is achieved by leveraging the semantic annotations of objects and image regions that have been created in the prior semantic segmentation research. 
  Second, it exploits visual saliency to determine the embedding locations within each semantic sensible region, which coincides with the fact that texts are often placed around homogeneous regions for better visibility in scenes.
  Third, it designs an adaptive text appearance model that determines the color and brightness of embedded texts by learning from the feature of real scene text images adaptively.
  The proposed technique has been evaluated over five public datasets and the experiments show its superior performance in training accurate and robust scene text detection and recognition models.
\keywords{Image synthesis, data augmentation, scene text detection, scene text recognition}
\end{abstract}

\section{Introduction}

  The capability of obtaining large amounts of annotated training images has become the bottleneck for effective and efficient development and deployment of deep neural networks (DNN) in various computer vision tasks. The current practice relies heavily on manual annotations, ranging from in-house annotation of small amounts of images to crowdsourcing based annotation of large amounts of images. On the other hand, the manual annotation approach is usually expensive, time-consuming, prone to human errors and difficult to scale while data are collected under different conditions or within different environments.

  Three approaches have been investigated to cope with the image annotation challenge in DNN training to the best of our knowledge. The first approach is probably the easiest and most widely adopted which augments training images by various label-preserving geometric transformations such as translation, rotation and flipping, as well as different intensity alternation operations such as blurring and histogram equalization \cite{geo-transform}. 
  The second approach is machine learning based which employs various semi-supervised and unsupervised learning techniques to create more annotated training images. For example, bootstrapping has been studied which combines the traditional self-training and co-training with the recent DNN training to search for more training samples from a large number of unannotated images \cite{Semi-Super,semi-super2}. In recent years, unsupervised DNN models such as Generative Adversarial Networks (GAN) \cite{GAN} have also been exploited to generate more annotated training images for DNN training \cite{gan2}.
  
  The third approach is image synthesis based which has been widely investigated in the area of computer graphics for the purpose of education, design simulation, advertising, entertainment, etc. \cite{img-syn}. It creates new images by modelling the physical behaviors of light and energy in combination of different rendering techniques 
such as embedding objects of interest (OOI) into a set of ``background images". To make the synthesized images useful for DNN training, the OOI should be embedded in the way that it looks as natural as possible. At the same time, sufficient variations should be included to ensure that the learned representation is broad enough to capture most possible OOI appearances in real scenes. 
  
  We propose a novel image synthesis technique that aims to create a large amount of annotated scene text images for training accurate and robust scene text detection and recognition models. The proposed technique consists of three innovative designs as listed:
\begin{description}
\item{1. It enables ``semantic coherent" image synthesis by embedding texts at semantically sensible regions within the background image as illustrated in Fig. 1, e.g. scene texts tend to appear over the wall or table surface instead of the food or plant leaves. We achieve the semantic coherence by leveraging the semantic annotations objects and image regions that have been created and are readily available in the semantic segmentation research, more details to be described in Section 3.1.}
\item{2. It exploits visual saliency to determine the embedding locations within each semantic coherent region as illustrated in Fig. 1. Specifically, texts are usually placed at homogeneous regions in scenes for better visibility and this can be perfectly captured using visual saliency. The exploitation of saliency guidance helps to synthesize more natural-looking scene text images, more details to be discussed in Section 3.2.}
\item{3. It designs a novel scene text appearance model that determines the color and brightness of source texts by learning from the feature of real scene text images adaptively. This is achieved by leveraging the similarity between the neighboring background of texts in scene images and the embedding locations within the background images, more details to be discussed in Section 3.3.}

\begin{figure}
\centering
\includegraphics[scale=0.435] {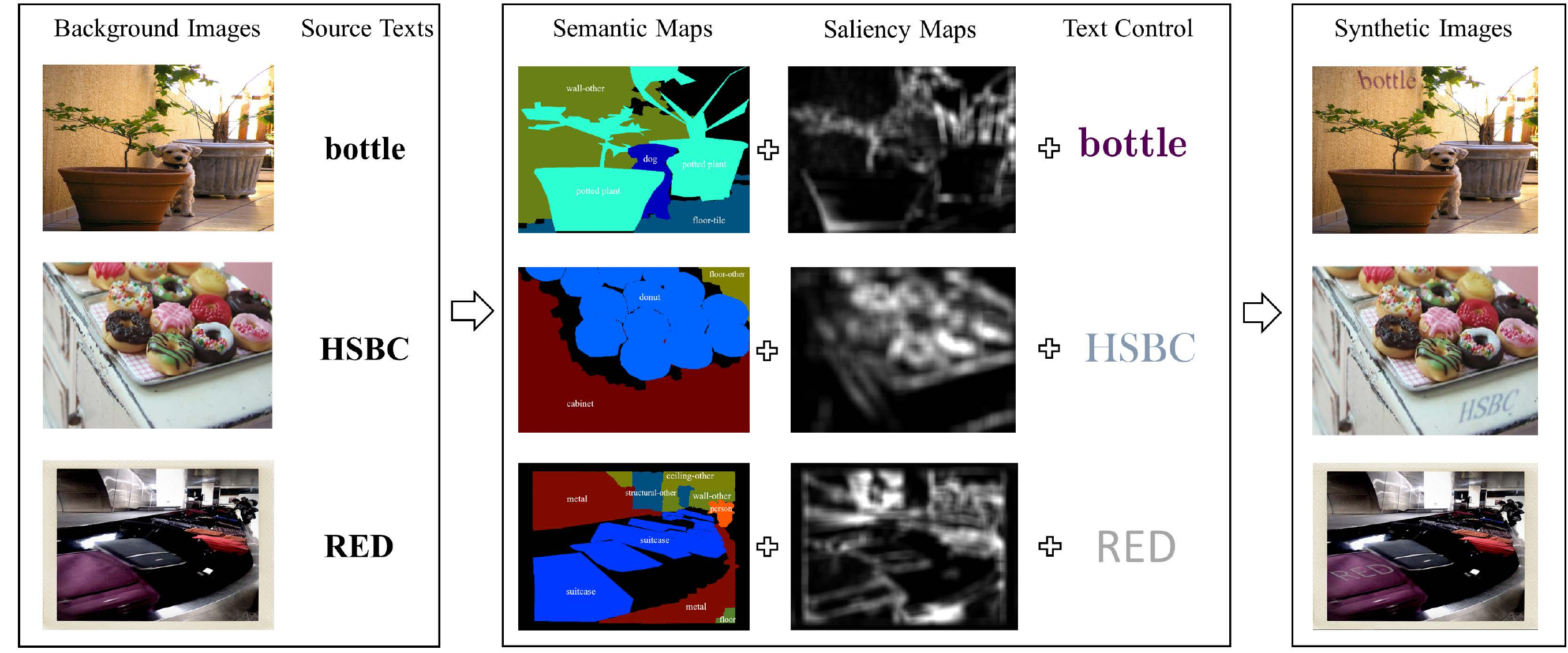}
\caption{The proposed scene text image synthesis technique: Given background images and source texts to be embedded into the background images as shown in the left-side box, a semantic map and a saliency map are first determined which are then combined to identify semantically sensible and apt locations for text embedding. The color, brightness, and orientation of the source texts are further determined adaptively according to the color, brightness, and contextual structures around the embedding locations within the background image. Pictures in the right-side box show scene text images synthesized by the proposed technique.}

\label{1}
\end{figure}
\end{description}

\section{Related Work}

\textbf{Image Synthesis} Photorealistically inserting objects into images has been studied extensively as one mean of image synthesis in the computer graphics research \cite{obj-syn}. The target is to achieve insertion verisimilitude, i.e., the true likeness of the synthesized images by controlling object size, object perspective (or orientation), environmental lighting, etc. For example, Karsch et al. \cite{render01} develop a semi-automatic technique that inserts objects into legacy photographs with photorealistic lighting and perspective.

In recent years, image synthesis has been investigated as a data augmentation approach for training accurate and robust DNN models when only a limited number of annotated images are available. For example, Jaderberg et al. \cite{Jaderberg14c} create a word generator and use the synthetic images to train text recognition networks. Dosovitskiy et al. \cite{Dosovitskiy} use synthetic floating chair images to train optical flow networks. Aldrian et al. \cite{Aldrian} propose an inverse rendering approach for synthesizing a 3D structure of faces. Yildirim et al. \cite{Yildirim} use the CNN features trained on synthetic faces to regress face pose parameters. Gupta el al. \cite{Gupta16} develop a fast and scalable engine to generate synthetic images of texts in scenes. On the other hand, most existing works do not fully consider semantic coherence, apt embedding locations and appearance of embedded objects which are critically important while applying the synthesized images to train DNN models.

\noindent
\textbf{Scene Text Detection } Scene text detection has been studied for years and it has attracted increasing interests in recent years as observed by a number of scene text reading competitions \cite{det1,det2,icdar13,det4}. Various detection techniques have been proposed from those using hand-crafted features and shallow models \cite{det6,det7,det11,det13,det16,det7,com-kang,shijianlu} to the recent efforts that design different DNN models to learn text features automatically \cite{det14,det15,det18,det19,Gupta16,det22,com-yin,com-zhang,det32,det19}. At the other end, different detection approaches have been explored including character-based systems \cite{det6,det11,det16,det14,det15,det18,det17} that first detect characters and then link up the detected characters into words or text lines, word-based systems \cite{Gupta16,det22,det24,det25,det26,det28,east} that treat words as objects for detection, and very recent line-based systems \cite{det19,det30} that treat text lines as objects for detection. Some other approaches \cite{det31,det32} localize multiple fine-scale text proposals and group them into text lines, which also show excellent performances.

On the other hand, scene text detection remains a very open research challenge. This can be observed from the limited scene text detection performance over those large-scale benchmarking datasets such as coco-text \cite{det33} and RCTW-17 dataset \cite{det1}, where the scene text detection performance is less affected by overfitting. One important factor that impedes the advance of the recent scene text detection research is very limited training data. In particular, the captured scene texts involve a tremendous amount of variation as texts may be printed in different fonts, colors and sizes and captured under different lightings, viewpoints, occlusion, background clutters, etc. A large amount of annotated scene text images are required to learn a comprehensive representation that captures the very different appearance of texts in scenes.

\noindent
\textbf{Scene Text Recognition} Scene text recognition has attracted increasing interests in recent years due to its numerous practical applications. Most existing systems aim to develop powerful character classifiers and some of them incorporate a language model, leading to state-of-the-art performance \cite{Jaderberg14c,rec5,com-abbyy,com-mishra,com-rodrguez,com-almazan,com-gordo,com-jaderberg,com-bissacco}. These systems perform character-level segmentation followed by character classification, and their performance is severely degraded by the character segmentation errors. Inspired by the great success of recurrent neural network (RNN) in handwriting recognition \cite{rec7}, RNN has been studied for scene text recognition which learns continuous sequential features from words or text lines without requiring character segmentation \cite{rec1,AB,CD,rec9}. On the other hand, most scene text image datasets such as ICDAR2013 \cite{icdar13} and ICDAR2015 \cite{det2} contain a few hundred/thousand training images only, which are too small to cover the very different text appearance in scenes.

\section{Scene Text Image Synthesis}
The proposed scene text image synthesis technique starts with two types of inputs including ``Background Images" and ``Source Texts" as illustrated in column 1 and 2 in Fig. 1. Given background images, the regions for text embedding can be determined by combining their ``Semantic Maps" and ``Saliency Maps" as illustrated in columns 3-4 in Fig. 1, where the ``Semantic Maps" are available as ground truth in the semantic image segmentation research and the ``Saliency Maps" can be determined using existing saliency models. The color and brightness of source texts can then be estimated adaptively according to the color and brightness of the determined text embedding regions as illustrated in column 5 in Fig. 1. Finally, ``Synthesized Images" are produced by placing the rendered texts at the embedding locations as illustrated in column 6 in Fig. 1.

\subsection{Semantic Coherence}

Semantic coherence (SC) refers to the target that texts should be embedded at semantically sensible regions within the background images. For example, texts should be placed over the fence boards instead of sky or sheep head where texts are rarely spotted in real scenes as illustrated in Fig. 2. The SC thus helps to create more semantically sensible foreground-background pairing which is very important to the visual representations as well as object detection and recognition models that are learned/trained by using the synthesized images. To the best of our knowledge, SC is largely neglected in earlier works that synthesize images for better deep network model training, e.g. the recent work \cite{Gupta16} that deals with a similar scene text image synthesis problem.
 
\begin{figure}
\centering
\begin{minipage}[t]{0.24\textwidth}
\includegraphics[width=2.9cm,height=2.5cm]{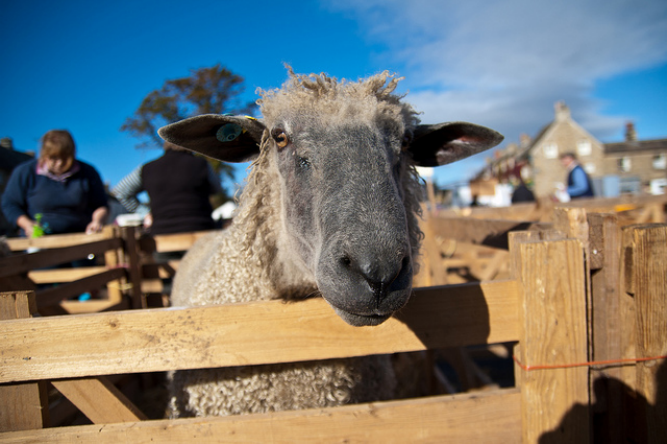}
\caption*{(a)}
\end{minipage}
\begin{minipage}[t]{0.24\textwidth}
\includegraphics[width=2.9cm,height=2.5cm]{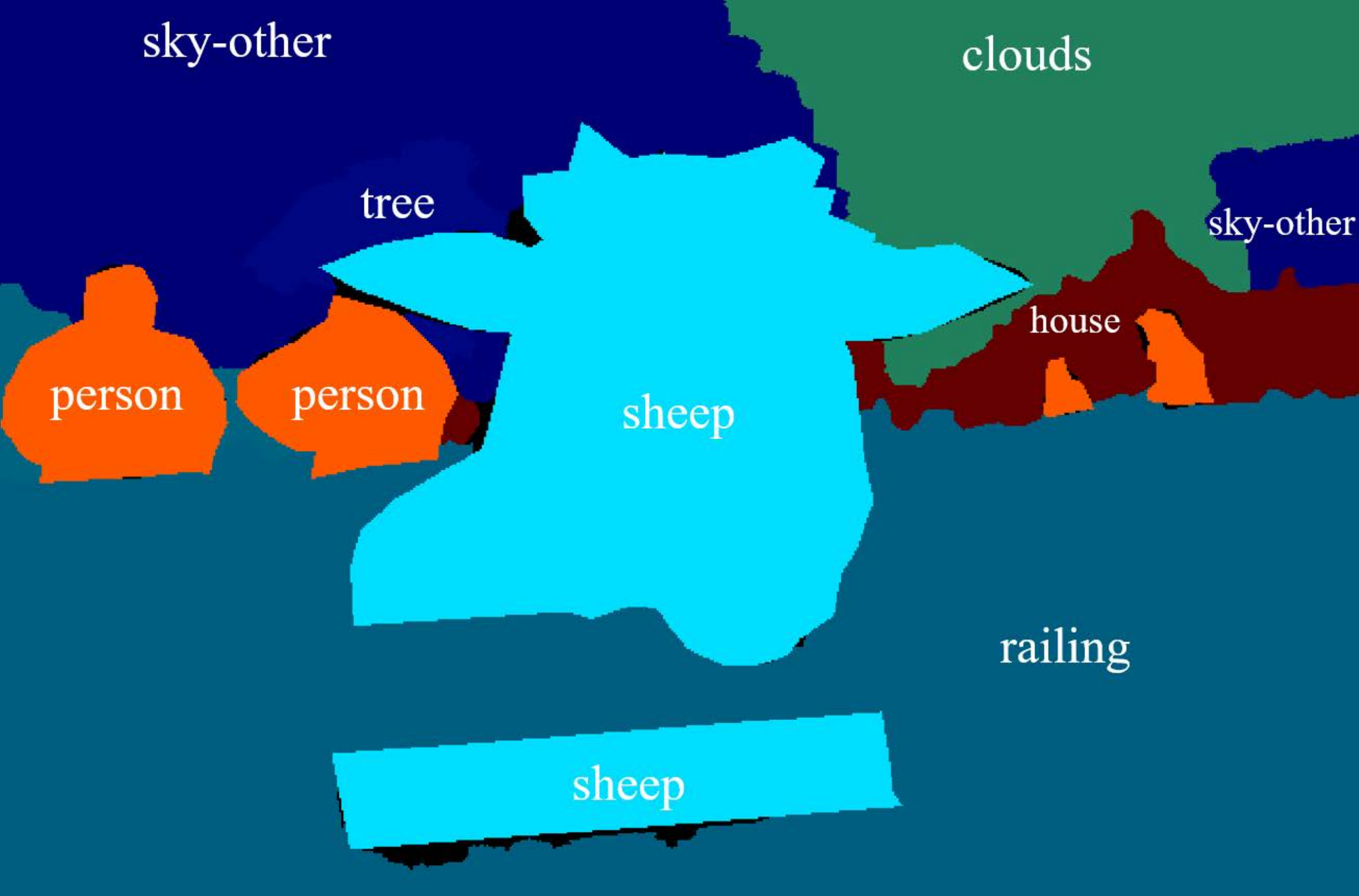}
\caption*{(b)}
\end{minipage}
\begin{minipage}[t]{0.24\textwidth}
\includegraphics[width=2.9cm,height=2.5cm]{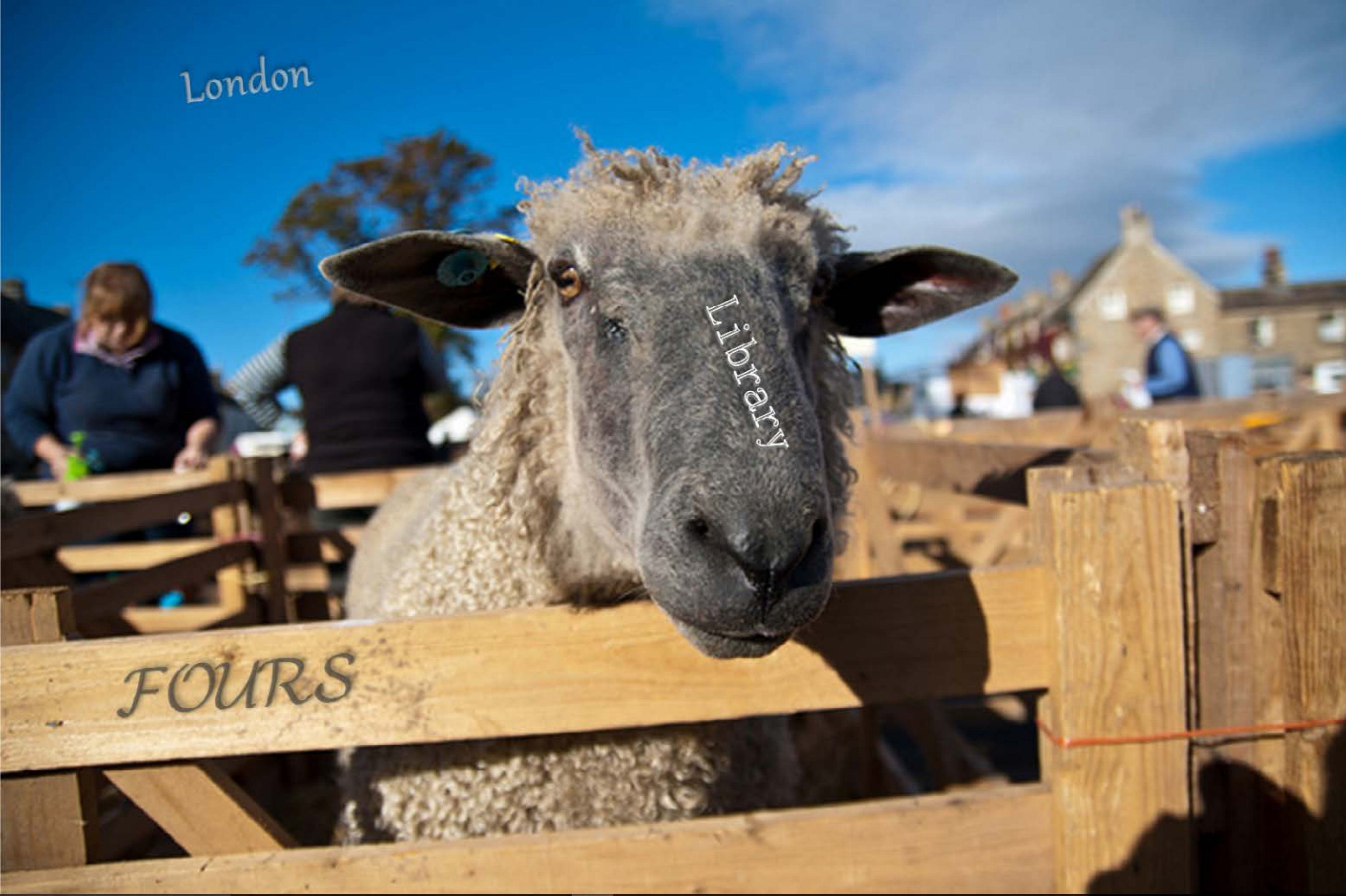}
\caption*{(c)}
\end{minipage}
\begin{minipage}[t]{0.24\textwidth}
\includegraphics[width=2.9cm,height=2.5cm]{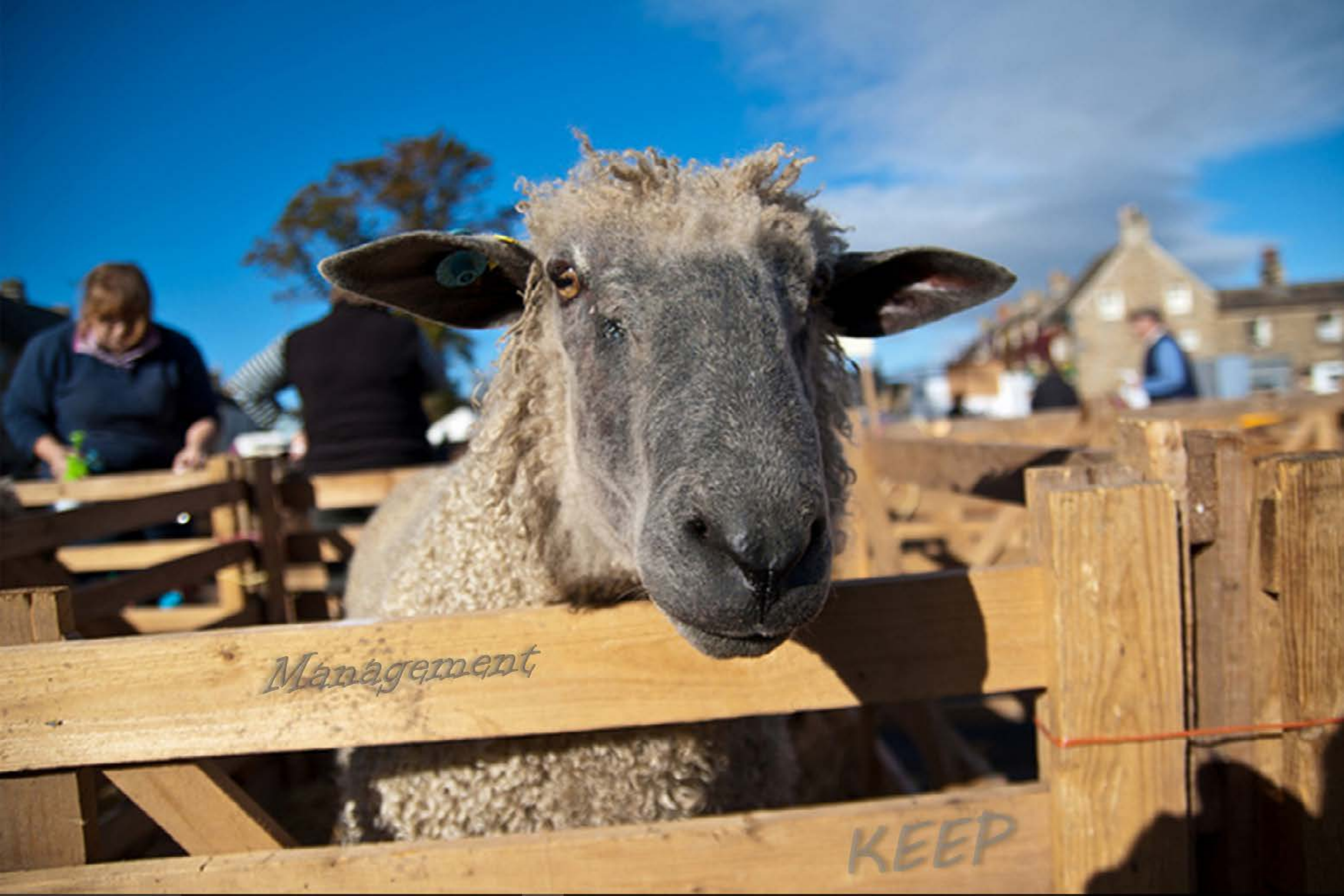}
\caption*{(d)}
\end{minipage}
\caption{Without semantic coherence (SC) as illustrated in (b), texts may be embedded at arbitrary regions such as sky and the head of sheep which are rarely spotted in scenes as illustrated in (c). SC helps to embed texts at semantically sensible regions as illustrated in (d).}
\end{figure}

We achieve the semantic coherence by exploiting a large amount of semantic annotations objects and image regions that have been created in the semantic image segmentation research. In particular, a number of semantic image segmentation datasets \cite{COCOStuff} have been created each of which comes with a set of ``ground truth" images that have been ``semantically" annotated. The ground truth annotation divides an image into a number of objects or regions at the pixel level where each object or region has a specific semantic annotation such as ``cloud", ``tree", ``person", ``sheep", etc. as illustrated in Fig.2.

To exploit SC for semantically sensible image synthesis, all available semantic annotations within the semantic segmentation datasets \cite{COCOStuff} are first classified into two lists where one list consists of objects or image regions that are semantically sensible for text embedding and the other consists of objects or image regions not semantically sensible for text embedding. Given some source texts for embedding and background images with region semantics, the image regions that are suitable for text embedding can thus be determined by checking through the pre-defined list of region semantics.

\subsection{Saliency Guidance}
Not every location within the semantically coherent objects or image regions are suitable for scene text embedding. For example, it's more suitable to embed scene texts over the surface of the yellow-color machine instead of across the two  neighboring surfaces as illustrated in Figs. 3c and 3d. Certain mechanisms are needed to further determine the exact scene text embedding locations within semantically coherent objects or image regions.

We exploit the human visual attention and scene text placement principle to determine the exact scene text embedding locations. To attract the human attention and eye balls, scene texts are usually placed around homogeneous regions such as signboards to create good contrast and visibility. With such observations, we make use of visual saliency as a guidance to determine the exact scene text embedding locations. In particular, homogeneous regions usually have lower saliency as compared with those highly contrasted and cluttered. Scene texts can thus be place at locations that have low saliency within the semantically coherent objects or image regions as described in the last subsection.

\begin{figure}
\centering
\begin{minipage}[t]{0.24\textwidth}
\includegraphics[width=2.9cm,height=2.5cm]{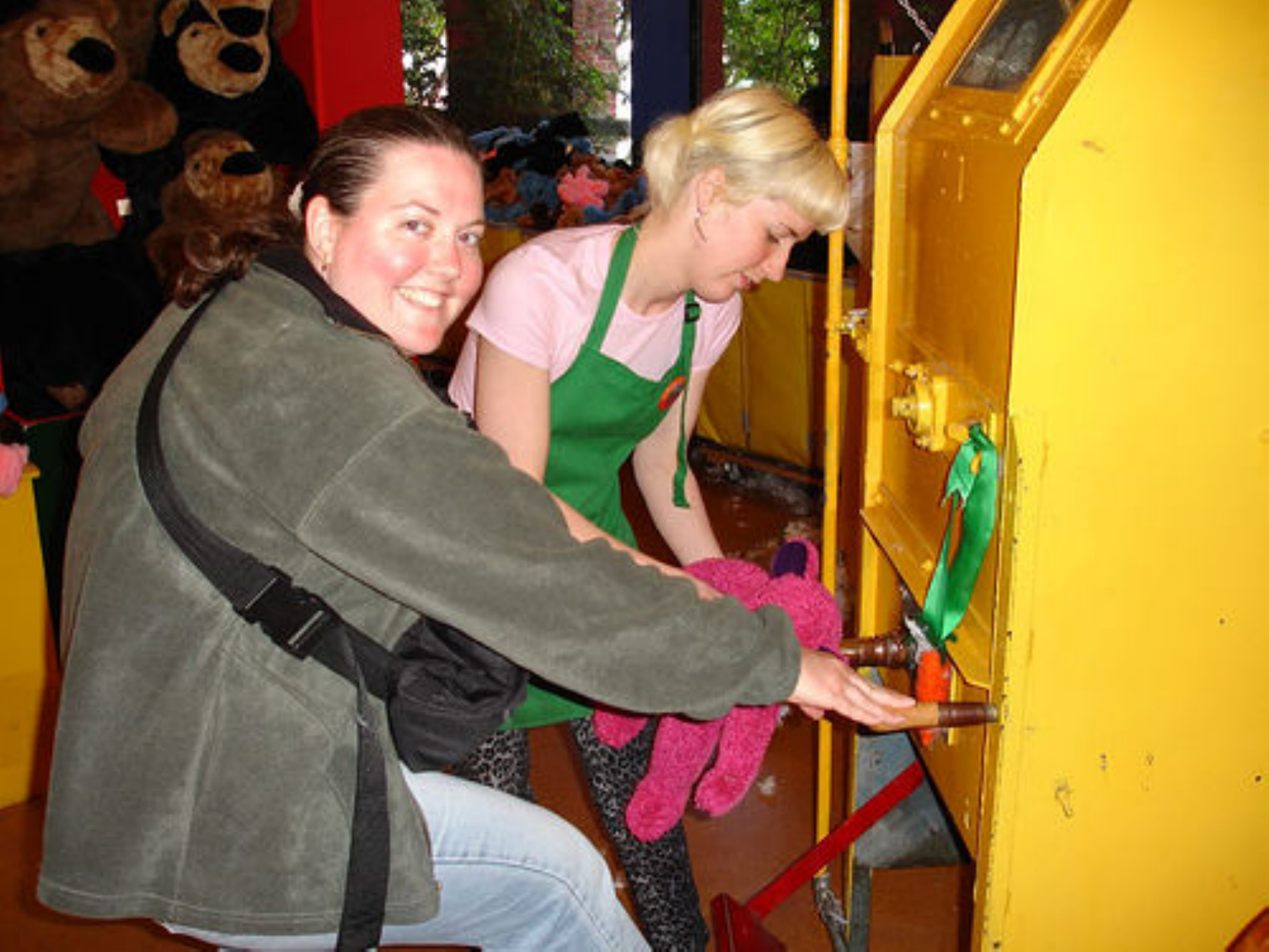}
\caption*{(a)}
\end{minipage}
\begin{minipage}[t]{0.24\textwidth}
\includegraphics[width=2.9cm,height=2.5cm]{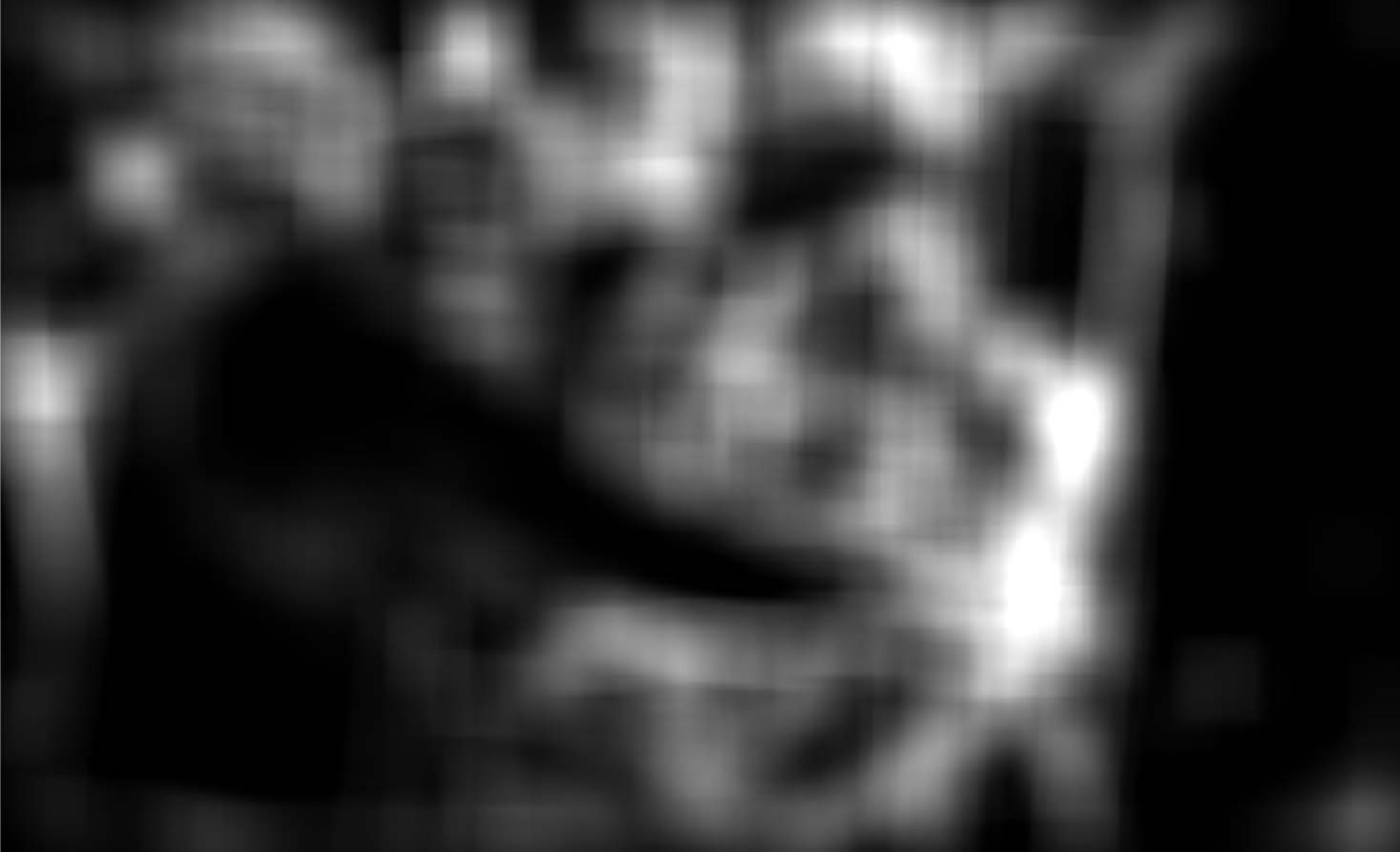}
\caption*{(b)}
\end{minipage}
\begin{minipage}[t]{0.24\textwidth}
\includegraphics[width=2.9cm,height=2.5cm]{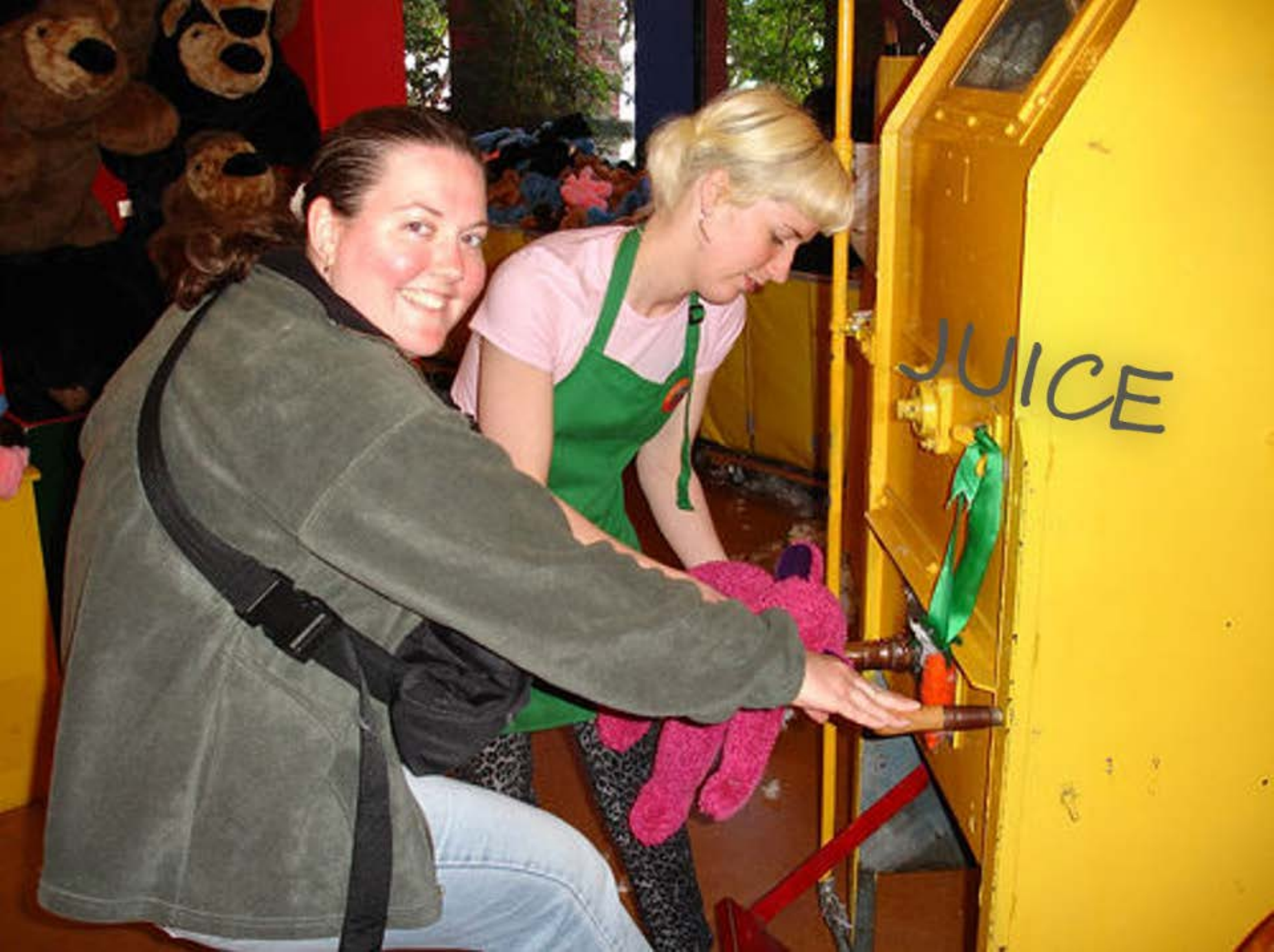}
\caption*{(c)}
\end{minipage}
\begin{minipage}[t]{0.24\textwidth}
\includegraphics[width=2.9cm,height=2.5cm]{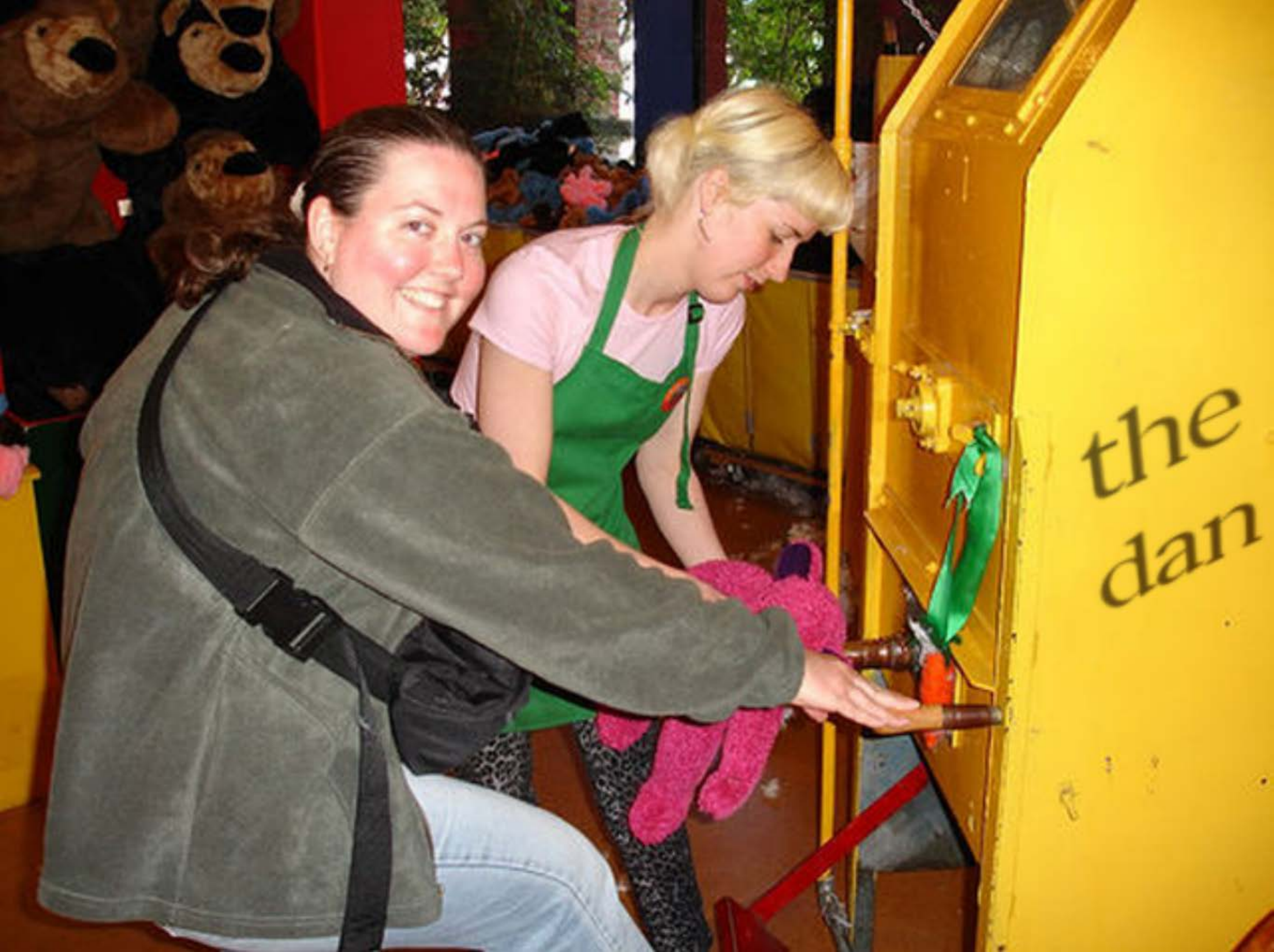}
\caption*{(d)}
\end{minipage}
\caption{Without saliency guidance (SG) as illustrated in (b), texts may be embedded across the object boundary as illustrated in (c) which are rarely spotted in scenes. SG thus helps to embed texts at right locations within the semantically sensible regions as illustrated in (d)}
\end{figure}

Quite a number of saliency models have been reported in the literature  \cite{saliency-1}. We adopt the saliency model in \cite{saliency-4} due to its good capture of local and global contrast. Given an image, the saliency model computes a saliency map as illustrated in Fig. 3, where homogeneous image regions usually have lower saliency. The locations that are suitable for text embedding can thus be determined by thresholding the computed saliency map. In our implemented system, a global threshold is used which is simply estimated by the mean of the computed saliency map. As Fig. 3 shows, the saliency guidance helps to embed texts at right locations within the semantically sensible regions. The use of saliency guidance further helps to improve the verisimilitude of the synthesized images as well as the learned visual representation of detection and recognition models.

\subsection{Adaptive Text Appearance}

Visual contrast as observed by low-level edges and corners is crucial feature while training object detection and recognition models. Texts in scenes are usually presented by linear strokes of different sizes and orientations which are rich in contrast-induced edges and corners. Effective control of the contrast between source texts and background images is thus very important to the usefulness of the synthesized images while applying them to train scene text detection and recognition models.

We design an adaptive contrast technique that controls the color and brightness of source texts according to what they look like in real scenes. The idea is to search for scene text image patches (readily available in a large amount of scene text annotations within existing datasets) whose background has similar color and brightness to the determined background regions as described in Sections 3.1 and 3.2. The color and brightness of the source texts can then be determined by referring to the color and brightness of text pixels within the searched scene text image patches.

The scene text image patches are derived from the scene text annotations as readily available in existing datasets such as ICDAR2013 \cite{icdar13}. For each text annotation, a HoG (histogram of oriented gradient) feature $H_b$ is first built by using the background region surrounding the text annotation under study. The mean and standard deviation of the color and brightness of the text pixels within the annotation box are also determined in the Lab color space, as denoted by ($\mu_L$, $\sigma_L$), ($\mu_a$, $\sigma_a$) and ($\mu_b$, $\sigma_b$). The background HoG $H_b$ and the text color and brightness statistics ($\mu_L$, $\sigma_L$), ($\mu_a$, $\sigma_a$) and ($\mu_b$, $\sigma_b$) of a large amount of scene text patches thus form a list of pairs as follows:
\begin{equation}
P = \big\{H_{b_1} : (\mu_{L_1}, \sigma_{L_1}, \mu_{a_1}, \sigma_{a_1}, \mu_{b_1}, \sigma_{b_1}), \cdots\, H_{b_i} : (\mu_{L_i}, \sigma_{L_i}, \mu_{a_i}, \sigma_{a_i}, \mu_{b_i}, \sigma_{b_i}), \cdots\big\}
\end{equation}


The $H_b$ in Eq. 1 will be used as the index of the annotated scene text image patch, and ($\mu_L$, $\sigma_L$), ($\mu_a$, $\sigma_a$) and ($\mu_b$, $\sigma_b$) will be used as a guidance to set the color and brightness of the source text. For each determined background patch (suitable for text embedding) as illustrated in Fig. 4, its HoG feature $H_s$ can be extracted and the scene text image patch that has the most similar background can thus be determined based on the similarity between $H_s$ and $H_b$. The color and brightness of the source text can thus be determined by taking the corresponding ($\mu_L$, $\mu_a$, $\mu_b$) plus random variations around ($\sigma_L$, $\sigma_a$, $\sigma_b$).

  The proposed technique also controls the orientation of the source texts adaptively according to certain contextual structures lying around the embedding locations within the background image. In particular, certain major structures (such as the table borders and the boundary between two connected wall surfaces as illustrated in Fig. 4) as well as their orientation can be estimated from the image gradient. The orientation of the source texts can then be determined by aligning with the major structures detected around the scene text embedding locations as illustrated in Fig. 4. Beyond the text alignment, the proposed technique also controls the font of the source texts by randomly selecting from a pre-defined font list as illustrated in Fig. 4.
  
\begin{figure}
\centering
\includegraphics[scale=0.575]{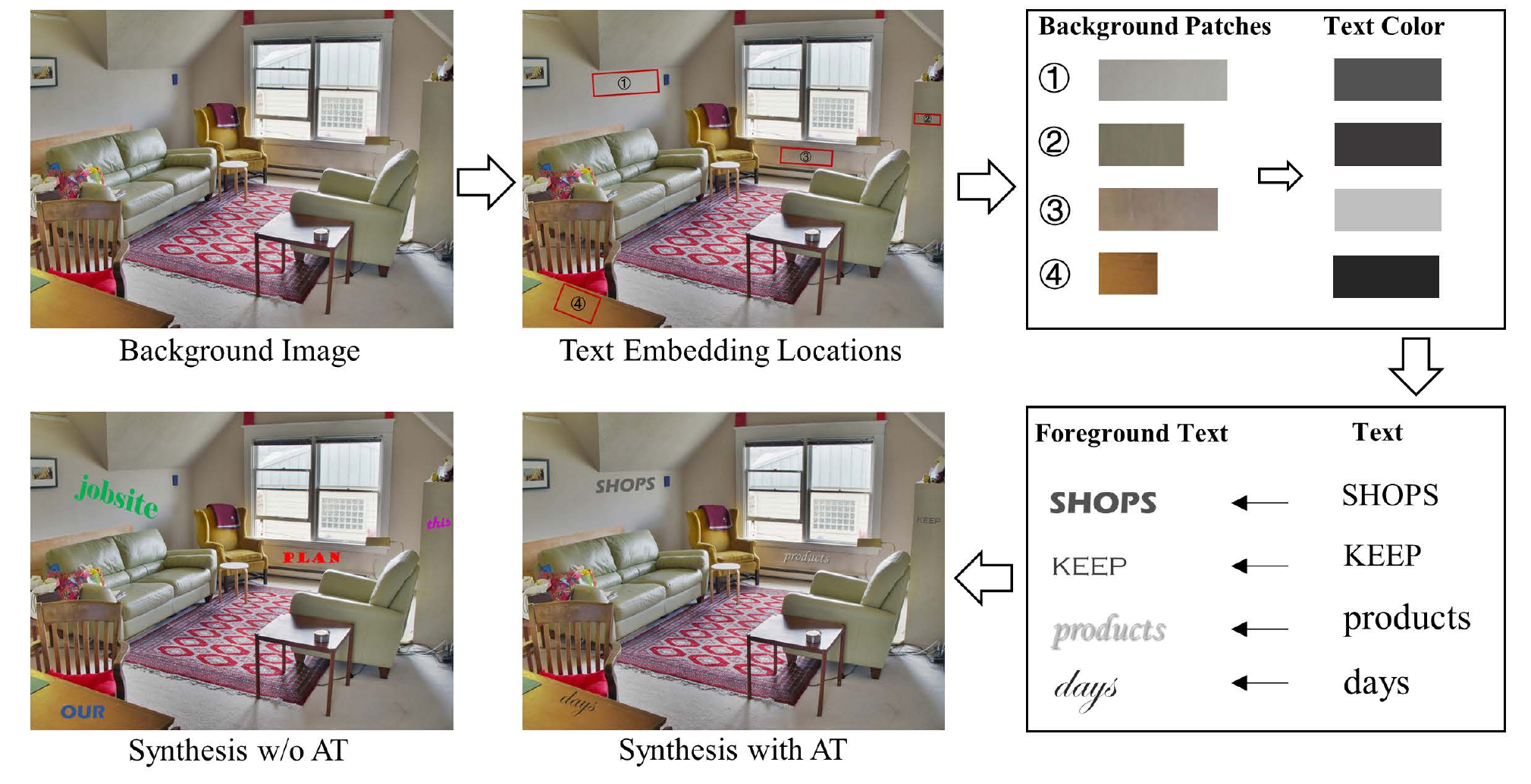}
\caption{Adaptive text appearance (ATA): The color and brightness of source texts are determined adaptively according to the color and brightness of the background image around the embedding locations as illustrated. The orientations of source texts are also adaptively determined according to the orientation of the contextual structures around the embedding locations. The ATA thus helps to produce more verisimilar text appearance as compared random setting of text color, brightness, and orientation.}
\label{1}
\end{figure}

\section{Implementations}

\subsection{Scene Text Detection}
We use an adapted version of EAST \cite{east} to train all scene text detection models to be discussed in Section 5.2. EAST is a simple but powerful detection model that yields fast and accurate scene text detection in scene images. The model directly predicts words or text lines of arbitrary orientations and quadrilateral shapes in the images. It utilizes the fully convolutional network (FCN) model that directly produces words or text-line level predictions, excluding unnecessary and redundant intermediate steps. Since the implementation of the original EAST is not available, we adopt an adapted implementation that uses ResNet-152 instead of PVANET \cite{pvanet} as the backbone network.

\subsection{Scene Text Recognition}
For the scene text recognition, we use the CRNN model \cite{rec1} to train all scene text recognition models to be described in Section 5.3. The CRNN model consists of the convolutional layers, the recurrent layers and a transcription layer which integrates feature extraction, sequence modelling and transcription into a unified framework. Different from most existing recognition models, the architecture in CRNN is end-to-end trainable and can handles sequences in arbitrary lengths, involving no character segmentation. Moreover, it is not confined to any predefined lexicon and can reach superior recognition performances in both lexicon-free and lexicon-based scene text recognition tasks.

\section{Experiments}
We evaluate the effectiveness of the proposed image synthesis technique on a scene text detection task and a scene text recognition task. The evaluations are performed over 5 public datasets to be discussed in the following subsections.

\subsection{Datasets and Evaluation Metrics}
The proposed technique is evaluated over five public datasets including ICDAR 2013 \cite{icdar13},ICDAR 2015 \cite{det2}, MSRA-TD500 \cite{det7}, IIIT5K \cite{iiit5k} and SVT\cite{com-abbyy}.

\textbf{ICDAR 2013} dataset is obtained from the Robust Reading Challenges 2013. It consists of 229 training images and 233 test images that capture text on sign boards, posters, etc. with word-level annotations. For recognition task, there are 848 word images for training recognition models and 1095 word images for recognition model evaluation. We use this dataset for both scene text detection and scene text recognition evaluations.

\textbf{ICDAR 2015} is a dataset of  incidental scene text and consists of 1,670 images (17,548 annotated text regions) acquired using the Google Glass. Incidental scene text refers to text that appears in the scene without the user taking any prior action in capturing. We use this dataset for the scene text detection evaluation.

\textbf{MSRA-TD500} dataset consists of 500 natural images (300 for training, 200 for test), which are taken from indoor and outdoor scenes using a pocket camera. The indoor images mainly capture signs, doorplates and caution plates while the outdoor images mostly capture guide boards and billboards with complex background. We use this dataset for the scene text detection evaluation.

\textbf{IIIT5K} dataset consists of 2000 training images and 3000 test images that are cropped from scene texts and born-digital images. For each image, there is a 50-word lexicon and a 1000-word lexicon. All lexicons consist of a ground truth word and some randomly picked words. We use this dataset for scene text recognition evaluation only.

\textbf{SVT} dataset consists of 249 street view images from which 647 words images are cropped. Each word image has a 50 word lexicon. We use this dataset for scene text recognition evaluation only.


For the scene text detection task, we use the evaluation algorithm
by Wolf et al. \cite{eval}. For the scene text recognition task, we perform evaluations based on the correctly recognized words (CRW) which can be calculated according to the ground truth transcription.

\begin{table*}[h]
\caption{Scene text detection recall (R), precision (P) and f-score (F) on the ICDAR2013, ICDAR2015 and MSRA-TD500 datasets, where ``\emph{EAST}" denotes the adapted EAST model as described in Section 4.1, ``Real" denotes the original training images within the respective datasets, ``Synth 1K" and ``Synth 10K" denote 1K and 10K synthesized images by our method.}
\centering 
\begin{tabular}{|p{3.75cm}|p{0.8cm}<{\centering} p{0.8cm}<{\centering} p{0.8cm}<{\centering}|p{0.8cm}<{\centering} p{0.8cm}<{\centering} p{0.8cm}<{\centering}|p{0.8cm}<{\centering} p{0.8cm}<{\centering} p{0.8cm}<{\centering}|}
\hline
Methods & \multicolumn{3}{c|}{ICDAR2013} & \multicolumn{3}{c|}{ICDAR2015} & \multicolumn{3}{c|}{MSRA-TD500}\\\hline 
 & R & P & F & R & P & F & R & P & F \\\hline
I2R NUS FAR \cite{icdar13} & 73.0 & 66.0 & 69.0 & - & - & - & - & - & -\\
\hline
TD-ICDAR \cite{det7} & - & - & - & - & - & - & 52.0 & 53.0 & 50.0\\
\hline
NJU \cite{det2} & - & - & - & 36.3 & 70.4 & 47.9 & - & - & -\\
\hline
Kang et al. \cite{com-kang} & - & - & - & - & - & - & 62.0 & 71.0 & 66.0\\
\hline
Yin et al. \cite{com-yin} & 65.1 & 84.0 & 73.4 & - & - & - & 63.0 & 81.0 & 71.0\\
\hline
Jaderberg et al. \cite{det22} & 68.0 & 86.7 & 76.2 & - & - & - & - & - & -\\
\hline
Zhang et al. \cite{com-zhang} & 78.0 & 88.0 & 83.0 & 43.1 & 70.8 & 53.6 & 67.0 & 83.0 & 74.0\\
\hline
Tian et al. \cite{det32} & 83.0 & 93.0 & 88.0 & 51.6 & 74.2 & 60.9 & - & - & -\\
\hline
Yao et al. \cite{det19} & 80.2 & 88.9 & 84.3 & 58.7 & 72.3 & 64.8 & {\bfseries 76.5} & 75.3 & 75.9\\
\hline
Gupta et al. \cite{Gupta16} & 76.4 & {\bfseries 93.8} & 84.2 & - & - & - & - & - & -\\
\hline
Zhou et al. \cite{east} & 82.7 & 92.6 & 87.4 & {\bfseries 78.3} & 83.3 & 80.7 & 67.4 & {\bfseries 87.3} & 76.1\\
\hline
\hline
\emph{EAST} (Real) & 80.5 & 85.6 & 83.0 & 75.8 & 84.1 & 79.7 & 69.2 & 78.1 & 73.4\\
\hline
\emph{EAST} (Real+Synth 1K) & 83.5 & 89.3 & 86.3 & 76.2 & 85.4 & 80.5 & 70.6 & 80.9 & 75.4\\
\hline
\emph{EAST} (Real+Synth 10K) & {\bfseries 85.0} & 91.7 & {\bfseries 88.3} & 77.2 & {\bfseries 87.1} & {\bfseries 81.9} & 72.7 & 85.7 & {\bfseries 78.6}\\\hline
\end{tabular}
\label{tab:PPer}
\end{table*}

\begin{table*}[h]
\caption{Scene text detection performance on the ICDAR2013 dataset by using the adapted EAST model as described in Section 4.1, where ``Synth" and ``Gupta" denote images synthesized by our method and Gupta et al. \cite{Gupta16} respectively, ``1K" and ``10K" denote the number of synthetic images used, ``Random" means embedding texts at random locations, SC, SG and ATA refer to semantic coherence, saliency guidance, and adaptive text appearance.}
\centering 
\begin{tabular}{l ccc} 
\hline 
Training Data & Recall & Precision & F-measure\\\hline 
ICDAR2013 (Baseline) & 80.49 & 85.56 & 82.95 \\[0.5ex]
\hline
ICDAR2013 + 1k Synth (Random) & 81.66 & 84.49 & 83.08 \\
ICDAR2013 + 1k Synth (SC) & 82.15 & 86.34 & 84.19 \\
ICDAR2013 + 1k Synth (SG) & 82.26 & 87.33 & 84.72 \\
ICDAR2013 + 1k Synth (ATA) & 81.90 & 84.95 & 83.40 \\
ICDAR2013 + 1k Synth (SC+SG) & 82.74 & 89.39 & 85.94 \\
ICDAR2013 + 1k Synth (SC+ATA) & 82.79 & 87.54 & 85.10 \\
ICDAR2013 + 1k Synth (SG+ATA) & 82.70 & 88.95 & 85.72 \\
ICDAR2013 + 1k Synth (SC+SG+ATA) & 83.46 & 89.34 & 86.29 \\
ICDAR2013 + 10k Synth (SC+SG+ATA) & 85.02 & 91.74 & 88.25 \\
\hline 
ICDAR2013 + 1k Gupta \cite{Gupta16} & 82.81 & 89.01 & 85.80 \\
ICDAR2013 + 10k Gupta \cite{Gupta16} & 84.93 & 90.74 & 87.74 \\
\hline
\end{tabular}
\label{tab:PPer}
\end{table*}

\subsection{Scene Text Detection}
For the scene text detection task, the proposed image synthesis technique is evaluated over three public datasets ICDAR2013, ICDAR2015 and MSRA-TD500. We synthesize images by catering to specific characteristics of training images within each dataset in term of text transcripts, text languages, text annotation methods, etc. Take the ICDAR2013 dataset as an example. The source texts are all in English and the embedding is at word level because almost all texts in the ICDAR2013 are in English and annotated at word level. For the MSRA-TD500, the source texts are instead in a mixture of English and Chinese and the embedding is at text line level because MSRA-TD500 contains both English and Chinese texts with text line level annotations. In addition, the source texts are a mixture of texts from the respective training images and publicly available corpses. The number of embedded words or text lines is limited at the maximum of 5 for each background image since we have sufficient background images with semantic segmentation.

Table 1 shows experimental results by using the adapted EAST (denoted by \emph{EAST}) model as described in Section 4.1. For each dataset, we train a baseline model ``\emph{EAST} (Real)" by using the original training images only as well as two augmented models ``\emph{EAST} (Real+Synth 1K)" and ``\emph{EAST} (Real+Synth 10K)" that further include 1K and 10K our synthesized images in training, respectively. As Table 1 shows, the scene text detection performance is improved consistently for all three datasets when synthesized images are included in training. In addition, the performance improvements become more significant when the number of synthesis images increases from 1K to 10K. In fact, the trained models outperform most state-of-the-art models when 10K synthesis images are used, and we can foresee further performance improvements when a larger amount of synthesis images are included in training. Furthermore, we observe that the performance improvements for the ICDAR2015 dataset are not as significant as the other two datasets. The major reason is that the ICDAR2015 images are videos frames as captured by Google glass cameras many of which suffer from motion and/or out-of-focus blur, whereas our image synthesis pipeline does not include image blurring function. We conjecture that the scene text detection models will perform better for the ICDAR2015 dataset if we incorporate the image blurring into the image synthesis pipeline.


In particular, a f-score of 83.0 is obtained for the ICDAR2013 dataset when the model is trained using the original training images. The f-score is improved to 86.2 when 1K synthetic images are included, and further to 88.3 when 10K synthetic images are included in training. Similar improvements are observed for the ICDAR2015 dataset, where the f-score is improved from the baseline 79.7 to 80.5 and 81.9 when 1K and 10K synthetic images are included in training. 
For the MSRA-TD500, a f-score of 73.4 is obtained when only the original 300 training images are used in model training. The f-score is improved to 75.4 and 78.6 respectively, when 1K and 10K synthetic images are included in training. This further verifies the effectiveness of the synthesized scene text images that are produced by our proposed technique.

We also perform ablation study of the three proposed image synthesis designs including semantic coherence (SC), saliency guidance (SG) and adaptive text appearance (ATA). Table 2 shows the experimental results over the ICDAR2013 dataset. As Table 2 shows, the inclusion of synthesized images (including random embedding in ``ICDAR2013 + 1k Synth (Random)") consistently improves the scene text detection performance as compared with the baseline model ``ICDAR2013 (Baseline)" that is trained by using the original training images only. In addition, the inclusion of either one of our three designs help to improve the scene text detection performance beyond the random embedding, where SG improves the most as followed by SC and AC. When all three designs are included, the f-score reaches 86.26 which is much higher than 83.09 by random embedding. Furthermore, the f-score reaches 88.25 when 10K synthesized images are included in training.
We also compared our synthesized images with those created by Gupta et al. \cite{Gupta16} as shown in Table 2, where the scene text detection models using our synthesized training images show superior performance consistently.

\subsection{Scene Text Recognition}

\begin{table*}[t]
\caption{Scene text recognition performance over the ICDAR2013, IIIT5K and SVT datasets, where ``50" and ``1K" in the second row denote the lexicon size and ``None" means no lexicon used. CRNN denotes the model as described in Section 4.2, ``Real" denote the original training images, ``Ours 5M", ``Jaderberg 5M" and ``Gupta 5M" denote the 5 million images synthesized by our method, Jaderberg et al. \cite{Jaderberg14c} and Gupta et al. \cite{Gupta16} respectively. }
\centering 
\begin{tabular}{lccccccccc} 
\hline 
Methods & \multicolumn{3}{c}{ICDAR2013} & \multicolumn{3}{c}{IIIT5K} & \multicolumn{3}{c}{SVT}\\\hline 
 &  & None &  & 50 & 1k & None & & 50 & None \\\hline
ABBYY \cite{com-abbyy} &  & - &  & 24.3 & - & - & & 35.0 & - \\
Mishra et al. \cite{com-mishra} &  & - &  & 64.1 & 57.5 & - & & 73.2 & - \\
Rodrguez-Serrano et al. \cite{com-rodrguez} &  & - &  & 76.1 & 57.4 & - & & 70.0 & - \\
Yao et al. \cite{rec5} &  & - &  & 80.2 & 69.3 & - & & 75.9 & - \\
Almaz´an et al. \cite{com-almazan} &  & - &  & 91.2 & 82.1 & - & & 74.3 & - \\
Gordo \cite{com-gordo} &  & - &  & 93.3 & 86.6 & - & & 91.8 & - \\
Jaderberg et al. \cite{com-jaderberg} &  & 81.8 &  & 95.5 & 89.6 & - & & 93.2 & 71.7 \\
Shi et al. \cite{rec1} &  & 86.7 &  & 97.6 & 94.4 & 78.2 & & 96.4 & 80.8 \\
Bissacco et al. \cite{com-bissacco} &  & 87.6 &  & - & - & - & & 90.4 & 78.0 \\
Shi et al. \cite{rec9} &  & {\bfseries 88.6} &  & 96.2 & 93.8 & {\bfseries 81.9} & & 95.5 & {\bfseries 81.9} \\
\hline
CRNN (Real) &  & 31.2 &  & 64.4 & 54.4 & 38.7 & & 62.1 & 35.5 \\
CRNN (Real+Jaderberg 5M \cite{Jaderberg14c}) &  & 85.6 &  & 97.1 & 93.2 & 77.1 & & 95.6 & 79.9 \\
CRNN (Real+Gupta 5M \cite{Gupta16}) &  & 86.4 &  & 96.7 & 92.4 & 76.0 & & 95.3 & 79.2 \\
CRNN (Real+Ours 5M) &  & 87.1 &  &{\bfseries 98.1} &{\bfseries 95.3} & 79.3 & &{\bfseries 96.7} & 81.5 \\\hline
\end{tabular}
\label{tab:PPer}
\end{table*}

For the scene text recognition task, the proposed image synthesis technique is evaluated over three public datasets ICDAR2013, IIIT5K and SVT as shown in Table 3 where the CRNN is used as the recognition model as described in Section 4.2. The baseline model ``CRNN (Real)" is trained by combining all annotated word images within the training images of the three datasets. As Table 3 shows, the baseline recognition accuracy is very low because the three datasets contain around 3100 word images only. As a comparison, the recognition model ``CRNN (Real+Ours 5M)" achieves state-of-the-art performance, where the 5 million word images are directly cropped from our synthesized scene text images as described in the last subsection. The significant recognition accuracy improvement demonstrates the effectiveness of the proposed scene text image synthesis technique.

In particular, the correctly recognized words (CRW) increases to 87.1\% for the ICDAR2013 dataset (without using lexicon) when 5 million synthetic images (synthesized by our proposed method) are included in training. This CRW is significantly higher than the baseline 31.2\% when only the original 3100 word images are used in training. For the IIIT5K, the CRW is increased to 79.3\% (no lexicon) when the same 5 million word images are included in training. The CRW is further improved to 95.3\% and 98.1\%, respectively, when the lexicon size is 1K and 50. Similar CRW improvements are also observed on the SVT dataset as shown in Table 3.

We also benchmark our synthesized images with those created by Jaderberg et al. \cite{Jaderberg14c} and Gupta et al. \cite{Gupta16}. In particular, we take the same amounts of synthesized images (5 million) and train the scene text recognition model ``CRNN (Real+Jaderberg 5M \cite{Jaderberg14c})" and ``CRNN (Real+Gupta 5M \cite{Gupta16})" by using the same CRNN network. As Table 3 shows, the model trained by using our synthesized images outperforms the models trained by using the ``Jaderberg 5M"  and ``Gupta 5M" across all three datasets.
Note that the model by Shi et al. \cite{rec1} achieves similar accuracy as the ``CRNN (Real+Ours 5M)", but it uses 8 million synthesized images as created by Jaderberg et al. \cite{Jaderberg14c}.



The superior scene text recognition accuracy as well as the significant improvement in the scene text detection task as described in the last subsection is largely due to the three novel image synthesis designs which help to generate verisimilar scene text images as illustrated in Fig. 5. As Fig. 5 shows, the proposed scene text image synthesis technique is capable of embedding source texts at semantically sensible and apt locations within the background image. At the same time, it is also capable of setting the color, brightness and orientation of the embedded texts adaptively according to the color, brightness, and contextual structures around the embedding locations within the background image.

\begin{figure}
\centering
\subfigure {\includegraphics[width=.23\linewidth]{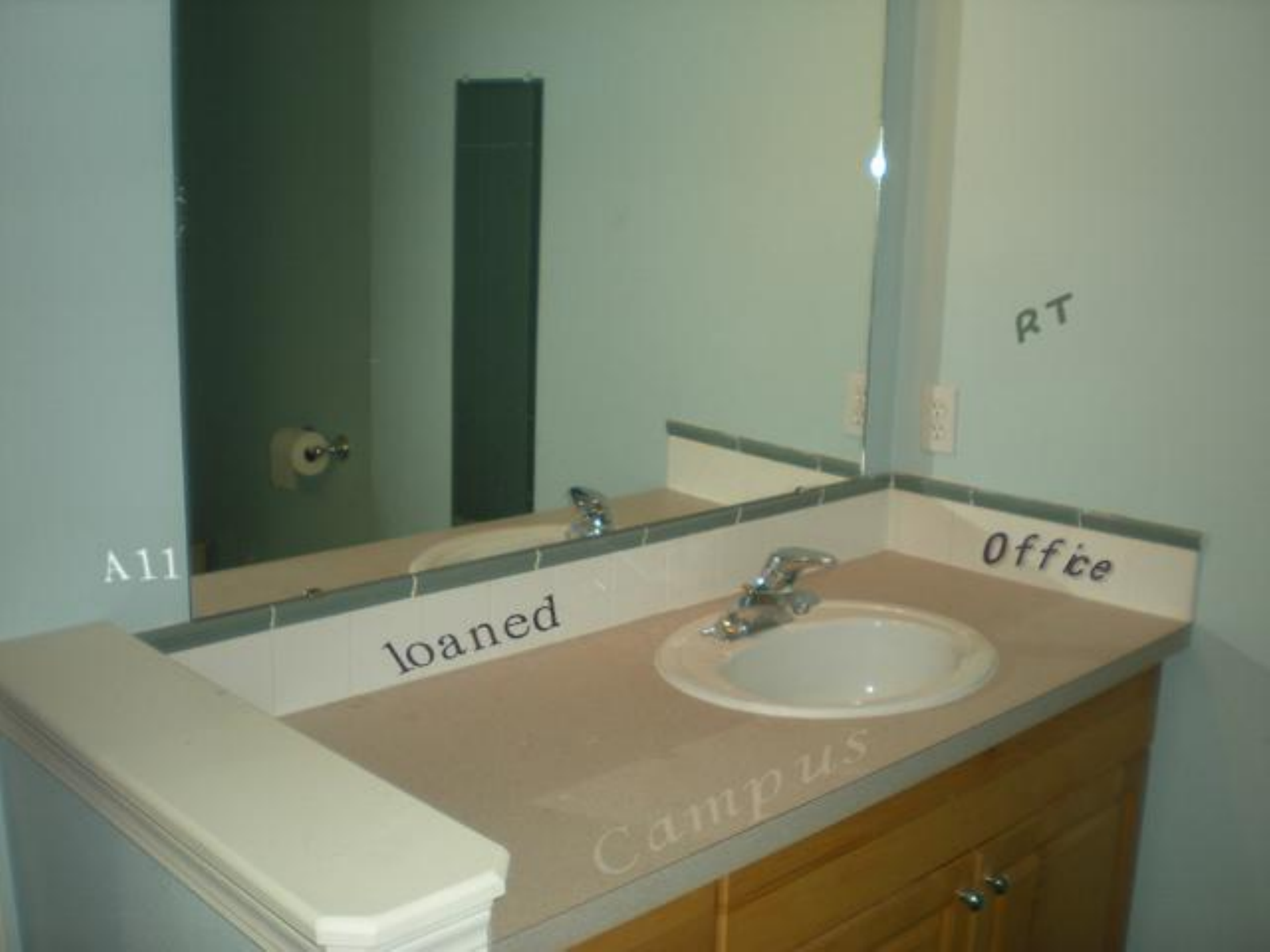}}
\subfigure {\includegraphics[width=.23\linewidth]{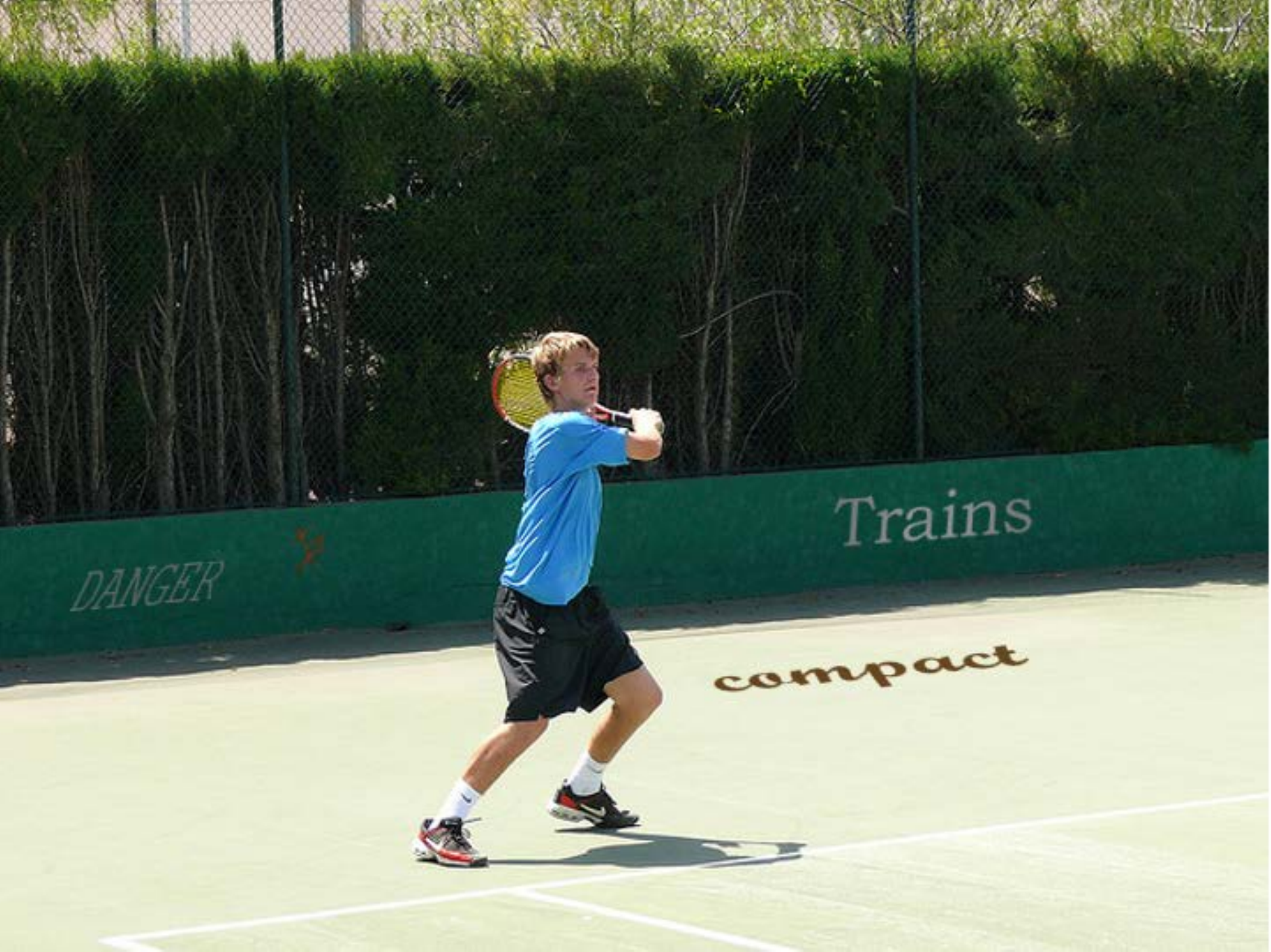}}
\subfigure {\includegraphics[width=.23\linewidth]{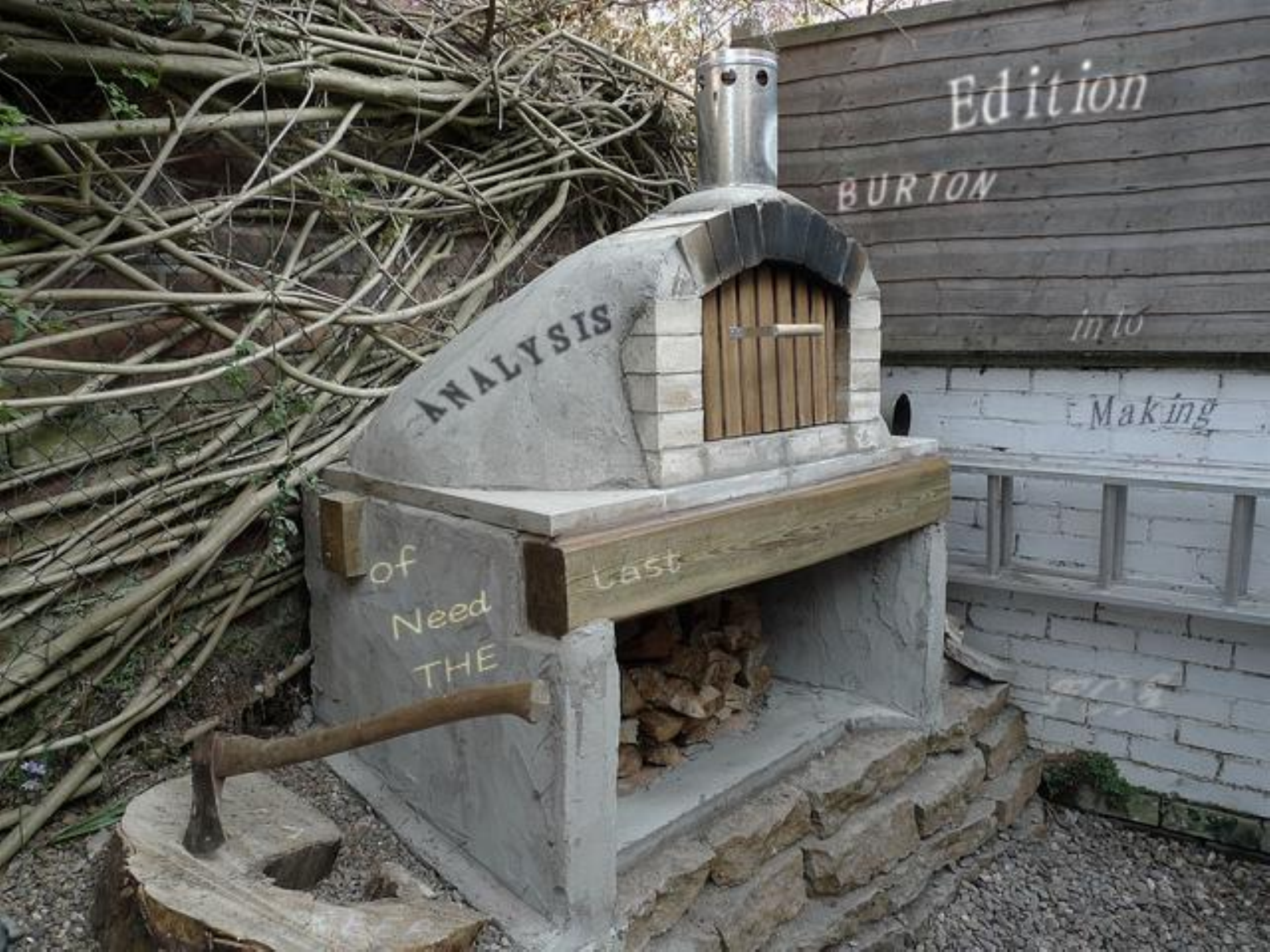}}
\subfigure {\includegraphics[width=.23\linewidth]{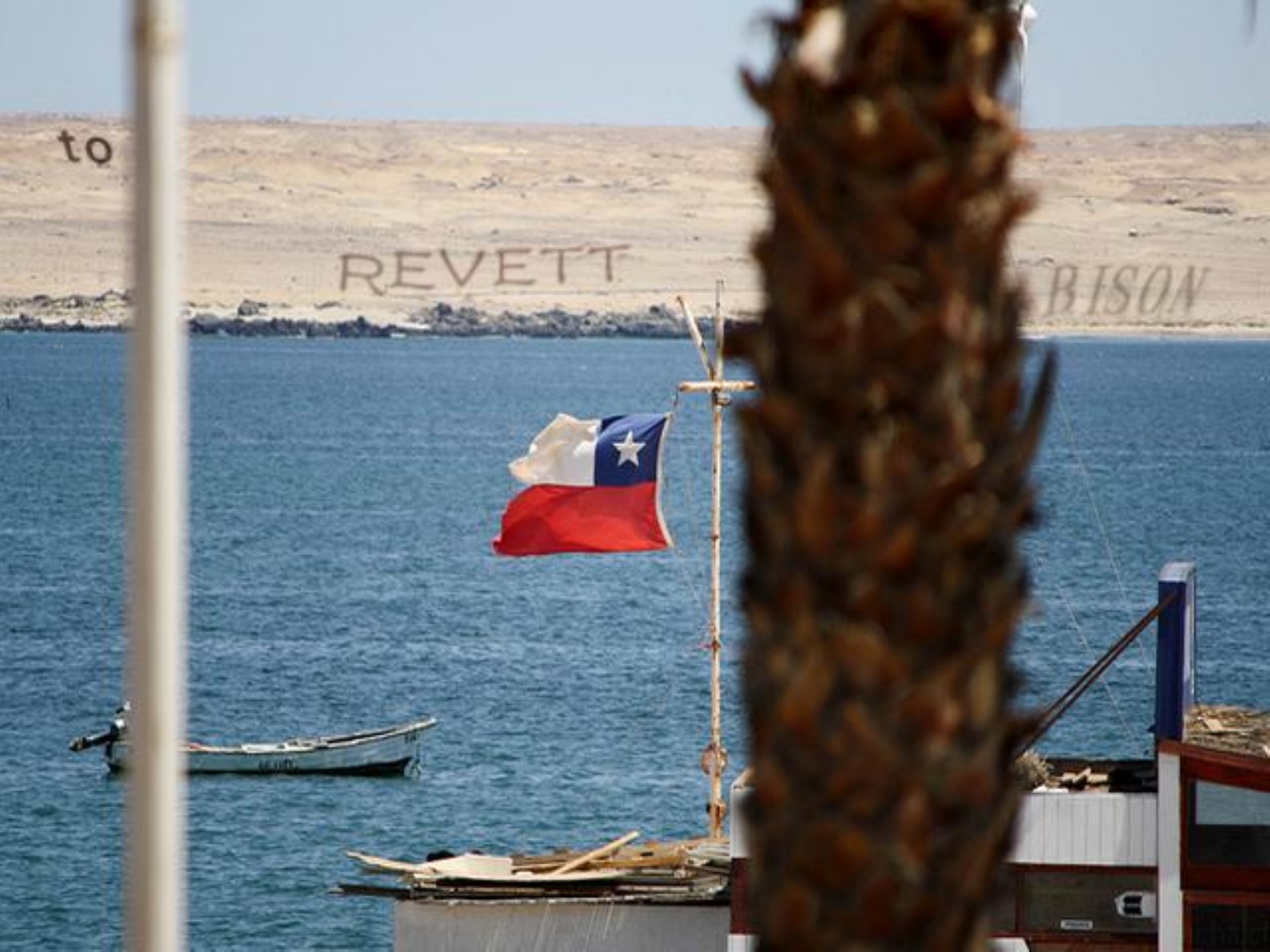}}

\subfigure {\includegraphics[width=.23\linewidth]{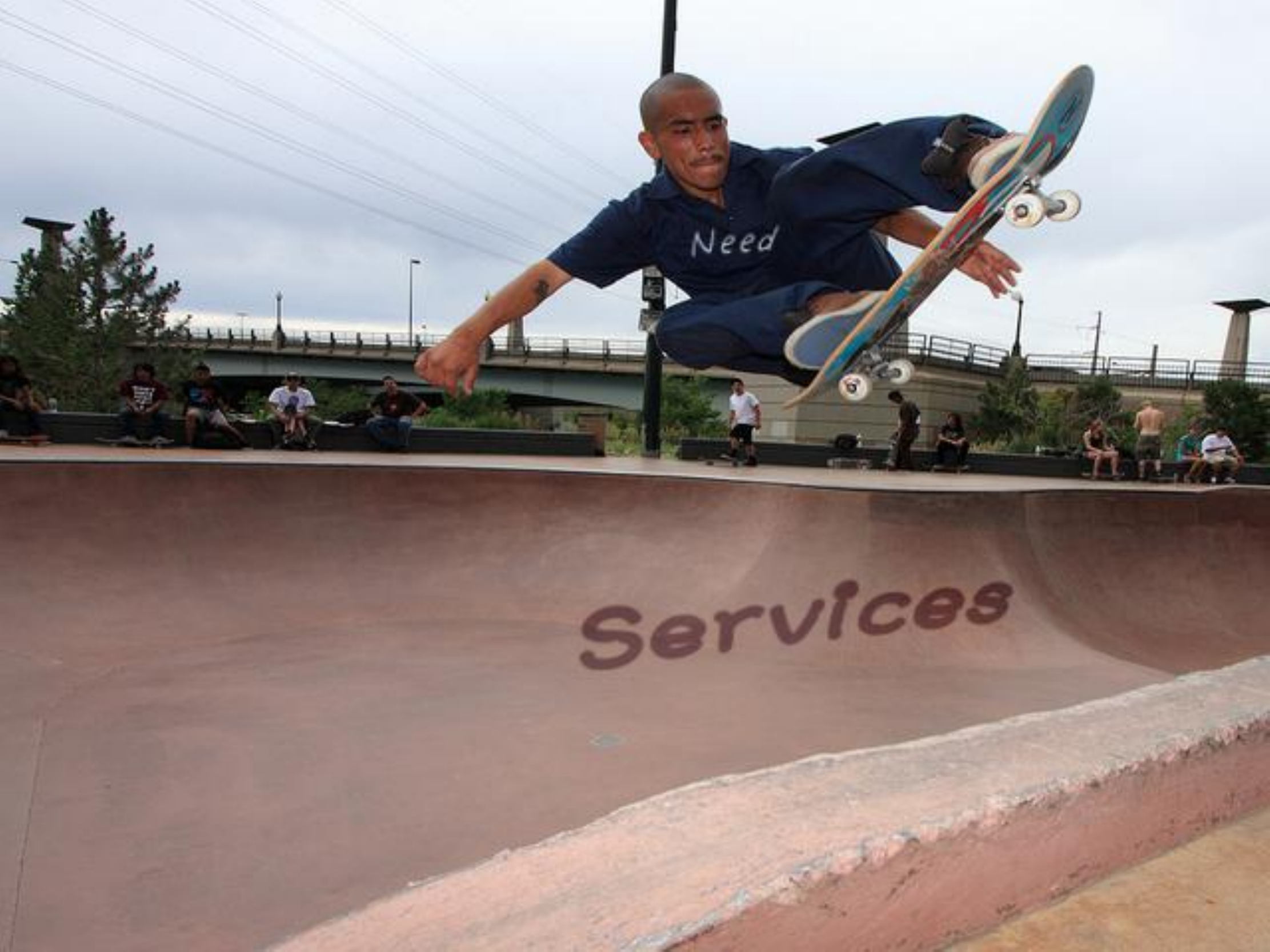}}
\subfigure {\includegraphics[width=.23\linewidth]{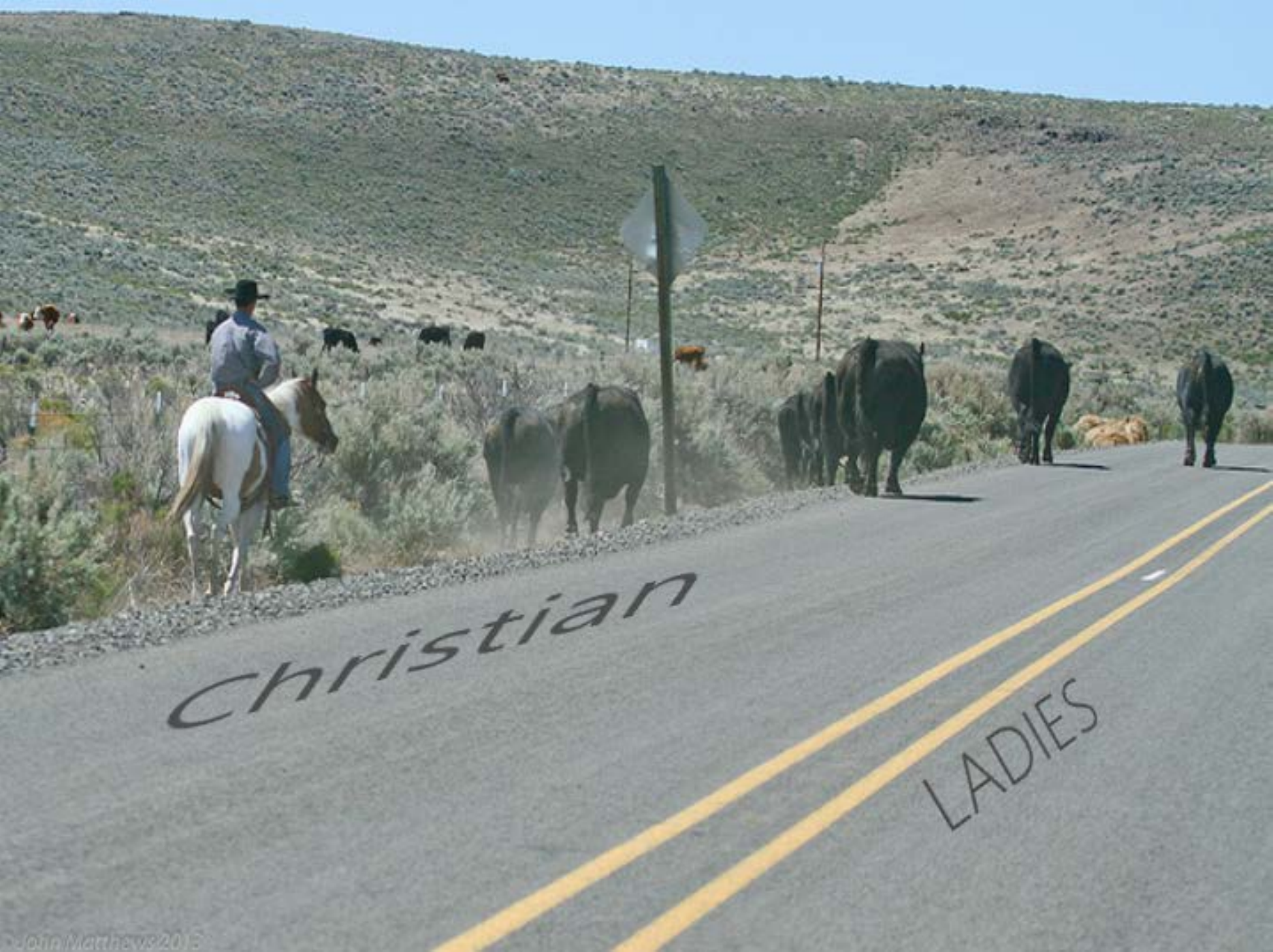}}
\subfigure {\includegraphics[width=.23\linewidth]{snipaste_20180117_145038.pdf}}
\subfigure {\includegraphics[width=.23\linewidth]{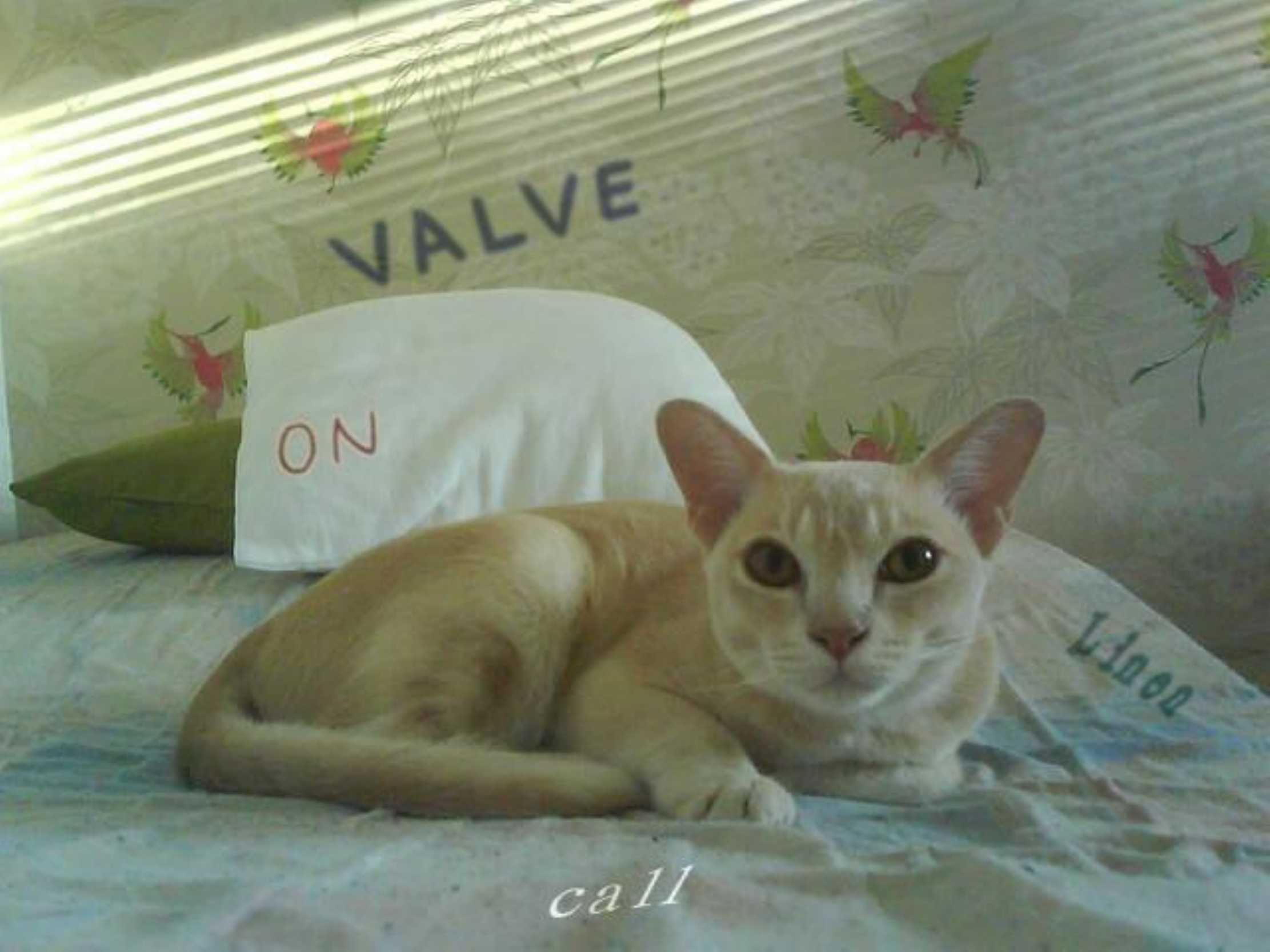}}

\subfigure {\includegraphics[width=.23\linewidth]{snipaste_20180314_125727.pdf}}
\subfigure {\includegraphics[width=.23\linewidth]{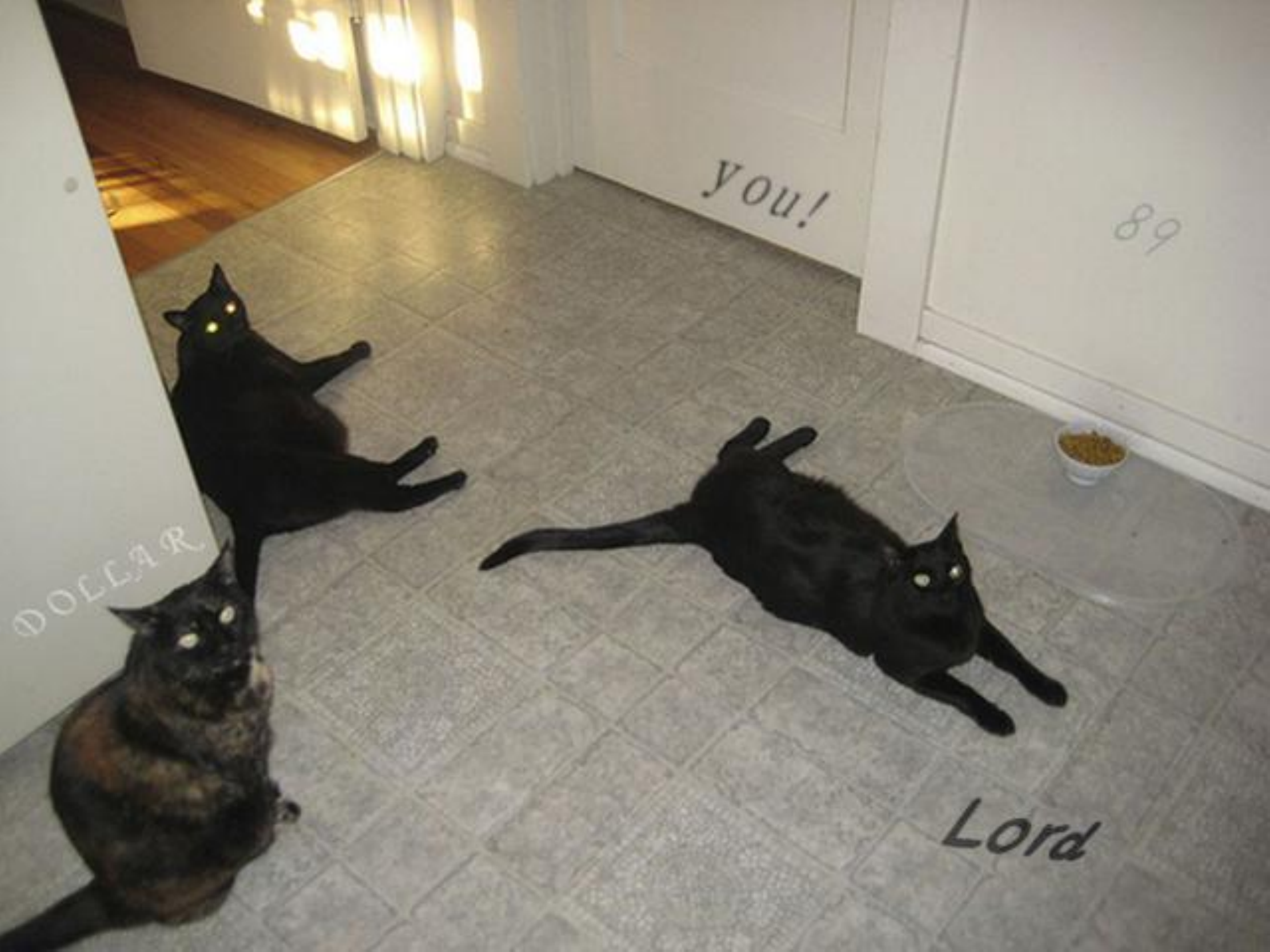}}
\subfigure {\includegraphics[width=.23\linewidth]{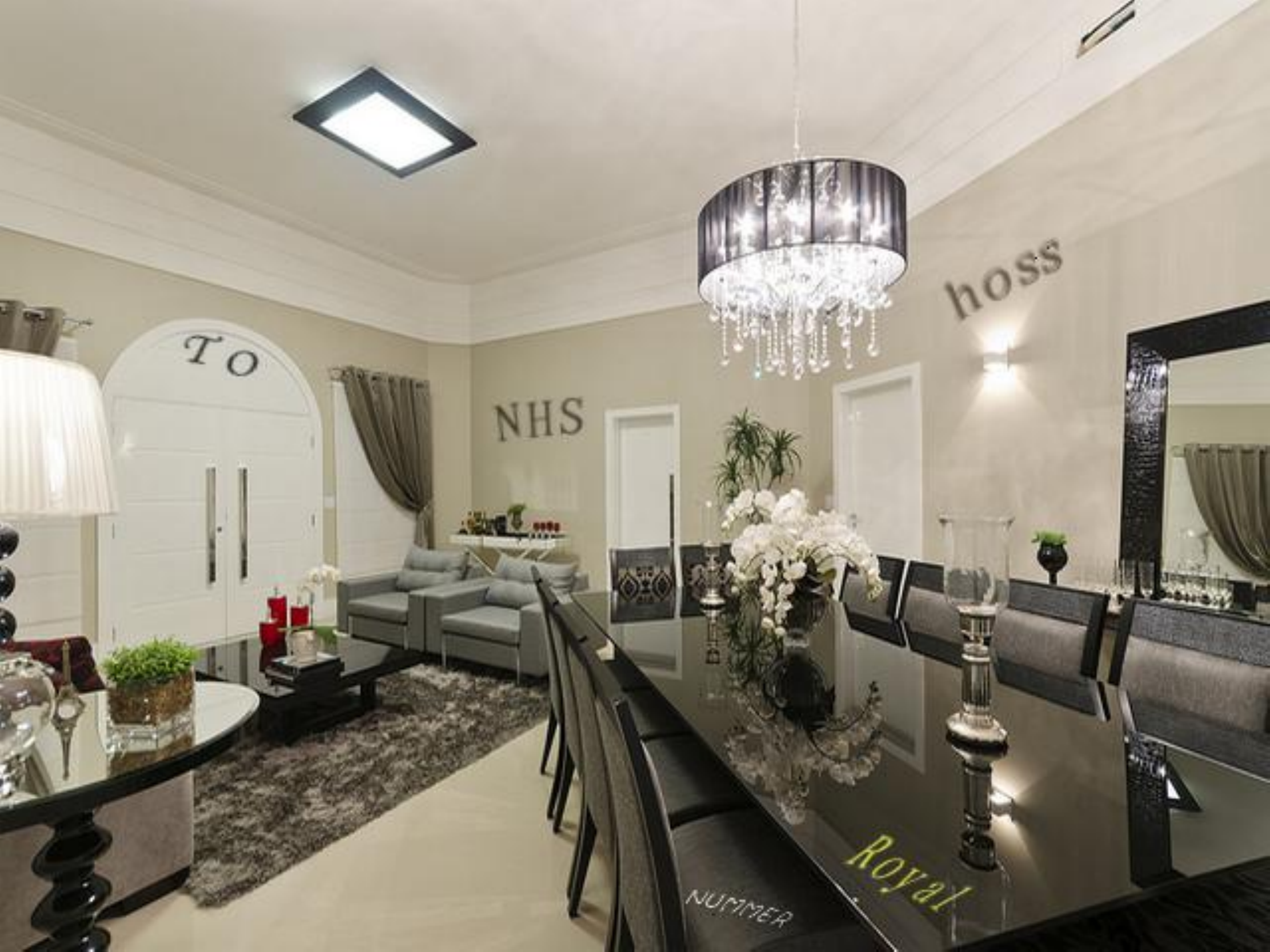}}
\subfigure {\includegraphics[width=.23\linewidth]{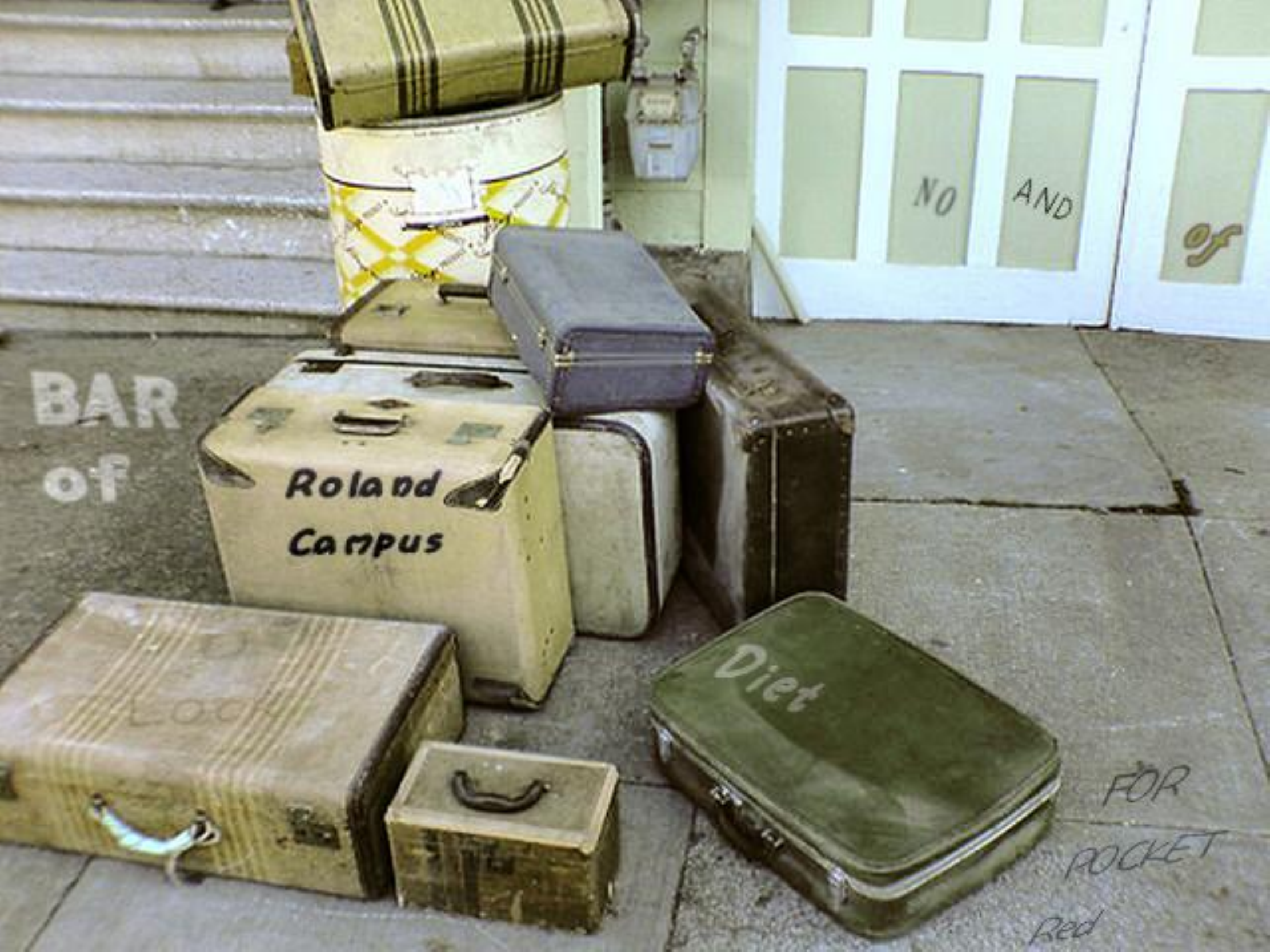}}

\caption{Several sample images from our synthesis dataset that show how the pro-posed semantic coherence, saliency guidance and adaptive text appearance work together for verisimilar text embedding in scene images automatically.}
\end{figure}

\section{Conclusions}
This paper presents a scene text image synthesis technique that aims to train accurate and robust scene text detection and recognition models. The proposed technique achieves verisimilar scene text image synthesis by combining three novel designs including semantic coherence, visual attention, and adaptive text appearance. Experiments over 5 public benchmarking datasets show that the proposed image synthesis technique helps to achieve state-of-the-art scene text detection and recognition performance.

A possible extension to our work is to further improve the appearance of source texts. We currently make use of the color and brightness statistics of real scene texts to guide the color and brightness of the embedded texts. The generated text appearance still has a gap as compared with the real scene texts because the color and brightness statistics do not capture the spatial distribution information. One possible improvement is to directly learn the text appearance of the dataset under study and use the learned model to determine the appearance of the source texts automatically.

\section{Acknowledgement}
This work is funded by the Ministry of Education, Singapore, under the project ``A semi-supervised learning approach for accurate and robust detection of texts in scenes" (RG128/17 (S)).

\bibliographystyle{splncs}
\bibliography{egbib}
\end{document}